\documentclass[lettersize,journal]{IEEEtran}
\usepackage{amsmath,amsfonts}
\usepackage{algorithmic}
\usepackage{algorithm}
\usepackage{array}
\usepackage{graphicx}
\usepackage{subcaption}
\usepackage{textcomp}
\usepackage{stfloats}
\usepackage{url}
\usepackage{verbatim}
\usepackage{graphicx}
\usepackage{cite}
\usepackage{booktabs} 
\usepackage{times}
\usepackage{soul}
\usepackage{url}
\usepackage{graphicx}
\usepackage[hidelinks]{hyperref}
\usepackage[utf8]{inputenc}
\usepackage{multirow}
\usepackage{booktabs}
\usepackage[normalem]{ulem}
\useunder{\uline}{\ul}{}
\begin{document}

\title{Hyperbolic Enhanced Representation Learning\\ for Incomplete Multi-view Clustering}

\author{Tianyi Chen$^{\dagger}$, Haobo Wang$^{\dagger}$, Kai Tang, Gengyu Lyu, Tianlei Hu, Gang Chen, Hong Ma$^{*}$, Meixiang Xiang$^{*}$
\thanks{Tianyi Chen and Haobo Wang contributed equally to this work.}        
\thanks{Tianyi Chen and Haobo Wang is with the School of Software Technology, Zhejiang University,
Zhejiang 310027, China (e-mail: tiannychen@zju.edu.cn; wanghaobo@zju.edu.cn).}
\thanks{Kai Tang, Tianlei Hu, Gang Chen are with the College of Computer
Science and Technology, Zhejiang University, Zhejiang 310027, China (e-mail:
kai.t@zju.edu.cn; htl@zju.edu.cn; cg@zju.edu.cn).}
\thanks{Gengyu Lyu is with the Beijing Key Laboratory of Traffic Data Analysis and Mining, Beijing Jiaotong
University, Beijing 100044, China, and also with the School of Computer and
Information Technology, Beijing Jiaotong University, Beijing 100044, China
(e-mail: lyugengyu@gmail.com).}
\thanks{Hong Ma and Meixiang Xiang is with the School of Medicine, Zhejiang University,
Zhejiang 310027, China (e-mail: hong\_ma@zju.edu.cn; xiangmx@zju.edu.cn).}
\thanks{Corresponding author: Hong Ma and Meixiang Xiang (e-mail: hong\_ma@zju.edu.cn; xiangmx@zju.edu.cn).}}

\markboth{Journal of \LaTeX\ Class Files,~Vol.~14, No.~8, August~2021}%
{Shell \MakeLowercase{\textit{et al.}}: A Sample Article Using IEEEtran.cls for IEEE Journals}


\maketitle


\begin{abstract}
Incomplete Multi-View Clustering (IMVC) aims to uncover latent cluster structures from partially observed multi-modal data, yet existing methods are predominantly developed in Euclidean space. Such flat geometry is often inadequate for modeling the hierarchical semantics of real-world data, especially under view incompleteness. As a result, learned representations tend to suffer from geometric mismatch and semantic blurring, where samples drift toward spatially close but semantically different neighbors, thereby degrading clustering discrimination and missing-view recovery. To address this limitation, we propose HERL, a Hyperbolic Enhanced Representation Learning framework for IMVC, which constructs a latent space in the Poincar'e ball. Specifically, HERL adopts a Euclidean affinity-guided contrastive backbone to learn cross-view representations and provide a basis for missing-view completion. Building upon this, we project features onto a constrained hyperbolic manifold and design a dual-constraint instance-level geometric alignment mechanism, which combines angular alignment to preserve semantic identity with distance-based anchoring to encode hierarchical compactness and refine structural relations. Furthermore, we introduce a hyperbolic prototype head to align cross-view hierarchy-aware prototype distributions, thereby mitigating structural drift and enforcing prototype-level semantic consistency. Through synergy between local geometric refinement and structural regularization, HERL enhances the discriminability, compactness, and recovery reliability of incomplete multi-view representations, leading to sharper cluster boundaries and more faithful imputation under missing data. Extensive experiments on four benchmark datasets demonstrate that HERL consistently outperforms state-of-the-art methods across a wide range of missing rates. Additional analyses and ablation studies verify that hyperbolic geometry provides a suitable inductive bias for capturing fine-grained semantic relations and latent hierarchies in incomplete multi-view clustering.
\end{abstract}


\section{Introduction}
Multi-view data are ubiquitous in real-world scenarios, where objects are described by heterogeneous modalities that theoretically provide complementary information for uncovering latent cluster structures (i.e., Multi-view Clustering)~\cite{zhang2025mixture}. In practice, however, the ideal assumption of view completeness rarely holds. Due to inevitable factors such as sensor failures in autonomous driving, privacy constraints in medical diagnosis, or varying acquisition costs in multimedia analysis, data are often partially observed~\cite{DBLP:journals/pami/LiuZLWTYSWG19}. This gives rise to Incomplete Multi-View Clustering (IMVC), a task that addresses the dual challenge of deriving discriminative cluster structures while maintaining robustness against incompleteness by maximizing the utility of fragmentary observations~\cite{liu2024sample,DBLP:conf/icml/YuC00Z25,liu2022localized,wang2024manifold}.

\begin{figure}[!t]
    \centering

    \begin{subfigure}[b]{\linewidth}
        \centering
        \includegraphics[width=0.9\linewidth]{intronew.pdf}
    \end{subfigure}
    
    \caption{Illustration of hierarchical structures in multi-view data and their geometric representations. Euclidean space provides a flat geometry that struggles to preserve hierarchical semantic relations. As an enhancement, hyperbolic space leverages its exponential volume to accommodate hierarchies, capturing fine-grained semantic distances.}
    \label{fig:intro}
\end{figure}

To this end, a large body of recent work formulates IMVC from a representation learning perspective, where contrastive learning has emerged as a central paradigm for enforcing cross-view semantic consistency. Some approaches \cite{SURE,DIVIDE} focus on refining instance-level contrastive relations to mitigate the influence of false negative pairs, while some other methods \cite{L2Dc,CPSPAN} further introduce category- or prototype-level contrastive objectives to promote semantically coherent clustering structures across views. Whether addressing missing data via explicit imputation \cite{ProImp} or learning view-invariant representations in an imputation-free manner \cite{I2MVC}, these methods converge on a common aim: learning discriminative representations from incomplete data to facilitate more accurate and robust downstream clustering.


Despite their prevalence, existing IMVC methods predominantly operate within Euclidean space, a setting we argue is geometrically suboptimal, particularly when distinctively modeling intrinsic structures from incomplete data. The flat geometry of Euclidean space lacks the capacity to accommodate the hierarchical semantic structures (e.g., fine-grained sub-categories) from the limited available complete samples, as shown in Fig.~\ref{fig:intro}. This geometric mismatch limits the representation quality: unable to discriminatively model latent hierarchies from the insufficient samples, the model struggles to differentiate between semantically similar clusters. Consequently, the boundaries between distinct yet related concepts become semantically blurred, obscuring the distinctiveness required for clustering. Furthermore, the absence of hierarchical constraints renders data recovery inherently ambiguous, causing representations to drift towards spatially proximal but semantically distinct neighbors.

To address these geometry limitations, we propose a Hyperbolic Enhanced Representation Learning framework for IMVC (\textbf{HERL}). Our key insight is that by constructing representations with explicit hierarchical structures, we can simultaneously enhance discrimination and constrain the recovery of missing data. Firstly, we employ Euclidean Affinity-Guided Contrastive Learning as the backbone to extract foundational representations and provide a preliminary basis for data imputation. Secondly, we project these embeddings onto a constrained Poincaré ball by applying feature clipping followed by the exponential map. Within this non-Euclidean space, we implement Hyperbolic Instance-level Geometric Alignment, which explicitly optimizes both angular and distance-based alignment to model intricate feature relationships and capture fine-grained hierarchies. Thirdly, to enforce global semantic structure, we design a hyperbolic prototype head that transforms embeddings into hierarchy-aware prototype distributions. This leverages hyperbolic geometry to enhance cross-view prototype-level semantic consistency and mitigate global structural drift. By unifying these objectives, HERL disentangles fine-grained semantic correlations to sharpen cluster boundaries and imposes geometric constraints to rectify the data recovery, yielding robust representation learning even under severe incompleteness. Extensive experiments on four benchmark datasets demonstrate that HERL consistently outperforms state-of-the-art approaches, with particularly notable accuracy improvements of \textbf{6.84\%} on Reuters at a missing rate of 0.3 and \textbf{4.07\%} on HandWritten at a missing rate of 0.1.

\section{Related Work}

\subsection{Incomplete Multi-view Clustering}

To cope with missing views, deep IMVC methods exploit representation learning at different levels to achieve robust clustering. Early works such as COMPLETER~\cite{COMPLETER} maximize cross-view mutual information to encourage consistent representations and enable view prediction. With the rise of contrastive learning, methods like DCP~\cite{DCP} combine instance-level contrastive objectives with clustering. However, as instance-wise learning suffers from false negative pairs, SURE~\cite{SURE} and DIVIDE~\cite{DIVIDE} introduce threshold-based or soft targets to improve robustness. Beyond sample-level alignment, recent studies explore prototype-level modeling for reliable semantics. ProImp~\cite{ProImp} and CPSPAN~\cite{CPSPAN} associate samples with class prototypes, enabling joint learning and stable imputation. Additionally, MVP~\cite{MVP} establishes inter-view correspondences in the VAE latent space. Other works bypass explicit imputation, such as ICMVC~\cite{ICMVC} (using GCN) and I2MVC~\cite{I2MVC} (using distillation). Despite their effectiveness, most approaches are formulated in Euclidean space, potentially suboptimal for complex hierarchical structures.

\subsection{Hyperbolic Learning}
Hyperbolic geometry has emerged as a powerful paradigm for representation learning, as its exponential volume growth naturally accommodates hierarchical structures where Euclidean spaces suffer from distortion~\cite{nickel2017poincare}. 
Initial works focused on embedding symbolic data like graphs and words~\cite{ganea2018hyperbolic}. 
Recently, this paradigm has been successfully extended to computer vision, demonstrating superior capability in capturing latent semantic hierarchies for image embeddings~\cite{klamt2019hyperbolic,ermolov2022hyperbolic} and contrastive learning~\cite{DBLP:conf/cvpr/GeMKL023}. 
Further studies broaden its use in specific tasks such as semantic segmentation~\cite{DBLP:conf/cvpr/AtighSANM22}, action understanding~\cite{franco2023hyperbolic}, and weakly supervised learning~\cite{DBLP:conf/nips/LengWTLG0024,liu2025hyperbolic}, underscoring its efficacy as a strong structural inductive bias.
In MVC, pioneering works like MHCN~\cite{MHCN} and CMHHC~\cite{CMHHC} integrate hyperbolic geometry but are restricted to complete views and instance-level alignment, overlooking data incompleteness and global consistency. Conversely, HERL is specifically tailored for IMVC, synergizing instance- and prototype-level constraints to ensure hierarchy-aware learning even with missing views.

\section{Preliminaries}

\subsection{Affinity-Guided Contrastive Objective}
\label{sec:preliminary_cl}
Standard contrastive learning treats all non-anchors as negatives, inevitably repelling semantically similar instances (the \textit{False Negative Problem}). To mitigate this, we adopt an affinity-guided paradigm~\cite{DIVIDE}. Instead of rigid binary labels, we utilize an affinity matrix $\mathbf{G}$ to provide soft supervision. Formally, the loss is defined as:
\begin{equation}
\label{eq:generic}
\begin{split}
    \mathcal{L}_{\text{con}}&(\mathbf{U}, \mathbf{V}; \mathbf{G}; \mathcal{S}) = \\
    &- \frac{1}{N} \sum_{i=1}^{N} \sum_{j=1}^{N} \mathbf{G}_{ij} \log \frac{\exp(\mathcal{S}(\mathbf{u}_i, \mathbf{v}_j) / \tau)}{\sum_{l=1}^{N} \exp(\mathcal{S}(\mathbf{u}_i, \mathbf{v}_l) / \tau)},
\end{split}
\end{equation}
where $\mathcal{S}(\cdot,\cdot)$ denotes the similarity metric (e.g., cosine similarity or hyperbolic distance) and $\tau$ is the temperature. This objective ensures representations respect the intrinsic data topology, pulling together semantically related samples even if they are not explicitly defined as anchor pairs.


\subsection{Hyperbolic Geometry}

Hyperbolic space $\mathbb{H}^n$ is a non-Euclidean Riemannian manifold characterized by constant negative curvature. A key property of this space is that its volume grows exponentially with the radius, making it intrinsically suitable for modeling hierarchical and tree-structured data where node capacity expands exponentially with depth~\cite{monath2019gradient}.

In this work, we represent hyperbolic space using the Poincaré ball model~\cite{nickel2017poincare}, which has been widely adopted in hyperbolic representation learning~\cite{liu2025hyperbolic}. The $n$-dimensional Poincar\'e ball is defined as $\mathbb{D}_c^n = \{ \mathbf{a} \in \mathbb{R}^n \mid c\|\mathbf{a}\|^2 < 1 \}$, where $c > 0$ is a hyperparameter that implies the curvature $K=-c$. The associated Riemannian metric takes the form $\mathcal{G}^{\mathbb{D}} = \lambda_c^2 \mathcal{G}^{\mathbb{E}}$, with $\lambda_c(\mathbf{x}) = \frac{2}{1 - c\|\mathbf{x}\|^2}$ denoting the conformal scaling factor and $\mathcal{G}^{\mathbb{E}}$ being the Euclidean metric tensor. As points approach the boundary of the ball, the scaling factor increases rapidly, leading to an exponential expansion of distances, enabling hyperbolic space to faithfully preserve hierarchical relations.
In order to project the feature $\mathbf{z}$ from Euclidean space $\mathbb{E}^n$ to hyperbolic space $\mathbb{H}^n$, we leverage the bijective exponential map at the origin with curvature $c$ formulated as:
\begin{equation}\label{eq:exp}
    \exp_\mathbf{o}^c(\mathbf{z}) = \tanh \left(\sqrt{c} \|\mathbf{z}\| \right) \frac{\mathbf{z}}{\sqrt{c}\|\mathbf{z}\|}.
\end{equation}
For the movement of points within the manifold, we utilize the {M\"{o}bius addition}~\cite{ganea2018hyperbolic}:
\begin{equation}
\label{equ:addition}
    \mathbf{x} \oplus_c \mathbf{y} = \frac{(1+2c\langle \mathbf{x}, \mathbf{y} \rangle + c\|\mathbf{y}\|^2) \mathbf{x}+ (1-c\|\mathbf{x}\|^2)\mathbf{y}}{1+2c\langle \mathbf{x}, \mathbf{y} \rangle + c^2 \|\mathbf{x}\|^2 \|\mathbf{y}\|^2}.
\end{equation}
To maintain numerical stability and prevent features from vanishing near the boundary,
we apply a feature clipping operator before the projection~\cite{guo2022clippedhyperbolicclassifierssuperhyperbolic}, defined as
$\mathcal{C}(\mathbf{z}) = \min \left\{1, \frac{cr}{\|\mathbf{z}\|} \right\} \cdot \mathbf{z}$,
where $cr$ is the clipping threshold.
 Consequently, the final hyperbolic mapping process employed in our framework is formally defined as:
\begin{equation}
\label{eq:final_mapping}
    \mathcal{H}(\mathbf{z}) = \exp_{\mathbf{o}}^c\!\left(\mathcal{C}(\mathbf{z})\right).
\end{equation}
This constrained projection guarantees that all embeddings strictly reside within the Poincaré ball, thereby ensuring valid geodesic computations in subsequent steps.

\section{Method}

\subsection{Backbone Network Construction}
\label{sec:overall_arch}
\begin{figure*}[!t]
    \centering

    \begin{subfigure}[b]{\linewidth}
        \centering
        \includegraphics[width=0.8\linewidth]{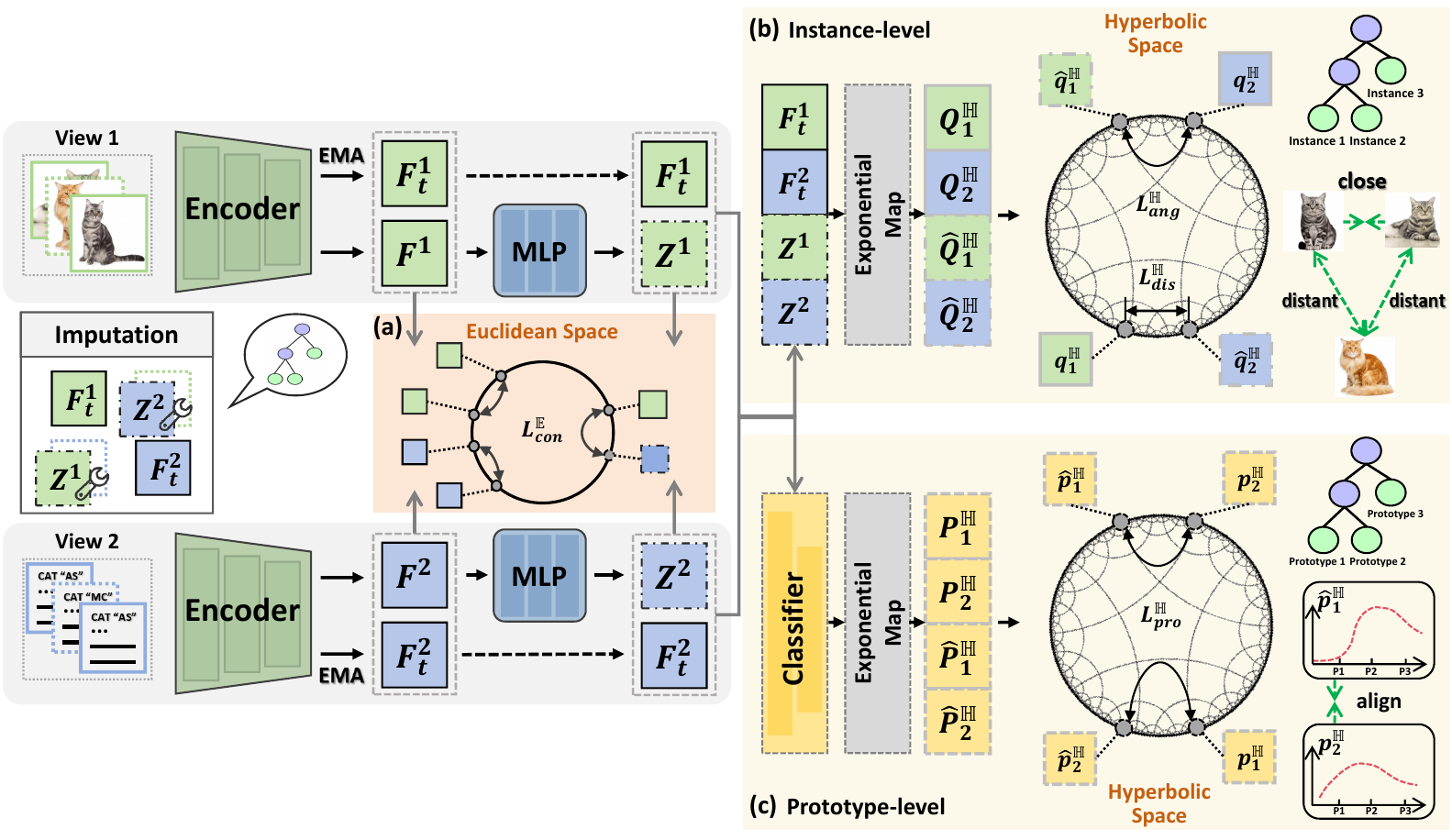}
    \end{subfigure}

    \caption{Overview of the {HERL} framework. 
    (a) {Euclidean Backbone} establishes basic representations to capture cross-view consistency while preserving view-specific information in flat space. 
    (b) {Hyperbolic Instance-level Geometric Alignment} projects features onto the Poincaré ball to capture fine-grained hierarchies via angular consistency and distance-based anchoring. 
    (c) {Hyperbolic Prototype-level Semantic Consistency} employs hyperbolic prototype heads to align features with hierarchy-aware prototype distributions.}

    \label{fig:method}
\end{figure*}

\noindent\textbf{Feature Extraction.} As illustrated in Fig.~\ref{fig:method}, HERL employs a dual-branch Euclidean backbone to extract view-specific student representations $\mathbf{F}^v$ and momentum-updated teacher counterparts $\mathbf{F}_t^v$. A projection head $g(\cdot)$ yields the predicted features $\mathbf{Z}^v = g(\mathbf{F}^v)$, establishing the preliminary Euclidean basis. To transcend the limitations of flat geometry and capture intrinsic hierarchies, we project these Euclidean features into the Poincaré ball manifold $\mathbb{D}_c^n$ via two parallel hyperbolic transformations to obtain instance-level and prototype-level embeddings, respectively:
\begin{equation} 
\label{eq:hyp_mapping} 
\begin{aligned} \mathbf{\hat Q}^{\mathbb{H}}_{v} &= \mathcal{H}(\mathbf{Z}^v), \quad 
\mathbf{\hat P}^{\mathbb{H}}_{v} = h^v(\mathbf{Z}^v). 
\end{aligned} 
\end{equation}
Here, the first set of embeddings, $\mathbf{\hat{Q}}^{\mathbb{H}}_v$ (and its teacher counterpart $\mathbf{Q}^{\mathbb{H}}_v = \mathcal{H}(\mathbf{F}^v_t)$), denotes the hyperbolic instance features utilized to capture fine-grained sample-to-sample geometric relationships. The second set, $\mathbf{\hat{P}}^{\mathbb{H}}_v$ (and $\mathbf{P}^{\mathbb{H}}_v = h^v(\mathbf{F}^v_t)$), represents the hierarchy-aware prototype distributions generated by view-specific hyperbolic prototype heads $h^v(\cdot)$, aiming to align global semantic structures across views. This dual-mapping strategy ensures the learned representations are both discriminative for instance alignment and robust for global semantic consistency.

\noindent\textbf{Euclidean Backbone Optimization.}
\label{sec:euclidean}
To establish a foundation for representation learning, we follow the strategy of DIVIDE~\cite{DIVIDE} to construct a high-order affinity graph $\mathbf{G}^v$ as a soft supervision signal.
Specifically, we derive a transition matrix $\mathbf{T}^v$ based on the pairwise Euclidean distances of teacher embeddings and perform a $t$-step random walk to capture global semantic relations (see supplemental material for full construction details):
\begin{equation}
\mathbf{G}^{v} = \xi \mathbb{I}_N + (1-\xi) (\mathbf{T}^{v})^t,
\end{equation}
where $\mathbb{I}_N$ is the identity matrix and $\xi$ is fixed to 0.5 to {balance the diffusion intensity}.
Adopting the paradigm defined in Eq.~\ref{eq:generic} and utilizing cosine similarity as the metric $\mathcal{S}$, the overall Euclidean objective is defined as:
\begin{equation}
\label{eq:con}
\mathcal{L}_{\text{con}}^{\mathbb{E}}\!=\!\sum_{v}\! \mathcal{L}_{\text{con}}(\mathbf{F}^v\!,\mathbf{F}_t^v;\mathbf{G}^{v}) + \!\sum_{v \neq u}\! \mathcal{L}_{\text{con}}(\mathbf{Z}^{v}\!,\mathbf{F}_t^u;\mathbf{G}^{u}),
\end{equation}
which jointly optimizes false-negative-aware view-specific structure and cross-view consistency.

\noindent\textbf{Data Recovery Mechanism.}
The projection head learned in the contrastive process can be interpreted as a semantics-preserving translator between views. It is therefore used to recover missing representations in incomplete data. Let $\mathbf{M} \in \{0,1\}^{N \times V}$ denote the observation mask, where $\mathbf{M}_{iv}=1$ indicates that sample $i$ is observed in view $v$. The unified completion rule is computed as:
\begin{equation}
\label{eq:completion}
    \tilde{\mathbf{Z}}^v = \mathbf{M}_{\cdot v} \odot \mathbf{F}_t^v + (\mathbf{1} - \mathbf{M}_{\cdot v}) \odot g(\mathbf{F}_t^u),
\end{equation}
where $\odot$ denotes the element-wise product. Subsequently, the completed representations from all views are concatenated and then processed by the $k$-means algorithm to produce the final clustering results.

While Euclidean geometry provides a preliminary basis for completion, the flat space is inherently limited in distinguishing semantically similar yet distinct representations. Consequently, the recovered features often suffer from semantic ambiguity and blurring, as evidenced by Fig.~\ref{fig:tsne}(a). This limitation necessitates a geometric upgrade to enforce clearer semantic boundaries for more reliable imputation.

\subsection{Hyperbolic Enhanced Representation Learning}


To address the geometric limitations of Euclidean space and further exploit intrinsic multi-level semantics, we introduce hyperbolic representation learning as a vital enhancement through two complementary components: an \textit{Hyperbolic Instance-level Geometric Alignment} module and a \textit{Prototype-level Semantic Consistency} module.
The former refines embeddings by optimizing both angular semantics and radial hierarchy within the Poincaré ball, while the latter enforces global structural consistency by aligning hierarchy-aware prototype distributions across views.

\subsubsection{Hyperbolic Instance-level Geometric Alignment}

This component {serves as a fine-grained geometric representation refiner} within the Poincaré ball, modeling {sample-to-sample relational relationships} to disentangle subtle semantic distinctions compressed in flat geometry. We achieve this by optimizing two complementary objectives: \textbf{Angular Alignment} and \textbf{Distance-based Anchoring}.

\noindent\textbf{Angular Alignment:} Given the hyperbolic representations $\mathbf{\hat Q}^{\mathbb{H}}_{v}$ and $\mathbf{Q}^{\mathbb{H}}_{v}$ (derived in Eq.~\eqref{eq:hyp_mapping}), we first prioritize {angular alignment}. Exploiting the conformal property of the Poincaré ball, where angles remain consistent with Euclidean geometry, we define the angular similarity as: $\mathcal{S}_{\text{a}}(\mathbf{a}, \mathbf{b}) = \frac{\mathbf{a} \cdot \mathbf{b}}{\|\mathbf{a}\| \, \|\mathbf{b}\|}$. Since this directional metric shares geometric consistency with the Euclidean backbone, we can effectively utilize the Euclidean affinity graph $\mathbf{G}^u$ as reliable semantic guidance:
\begin{equation}
\label{eq:ang}
    \mathcal{L}_{\text{ang}}^{\mathbb{H}} = \sum_{v \neq u} \mathcal{L}_{\text{con}}\Big(\mathbf{\hat Q}^{\mathbb{H}}_{v}, \mathbf{Q}^{\mathbb{H}}_{u}; \mathbf{G}^{u}; \mathcal{S}_{\text{a}}(\mathbf{\hat Q}^{\mathbb{H}}_{v}, \mathbf{Q}^{\mathbb{H}}_{u})\Big).
\end{equation}
\textit{Geometric Insight:} This objective functions as a \textit{semantic stabilizer}. By strictly aligning the limited available complete views, it partitions the Poincaré ball into distinct {semantic cones} radiating from the origin. This benefits both clustering and recovery: for representation learning, it ensures that instances are robustly separated by category direction; for data recovery, it constrains the completed views to fall within these correct semantic sectors. Even if the imputed features contain noise, this angular constraint prevents catastrophic category shifts, ensuring the recovered data serves the downstream clustering task effectively.

\noindent\textbf{Distance-based Anchoring:} Second, to capture the intrinsic data topology, we incorporate {distance-based anchoring}. We define the distance similarity as $\mathcal{S}_{\text{d}}(\mathbf{a}, \mathbf{b}) = -\mathcal{D}_{\mathbb{H}}(\mathbf{a}, \mathbf{b})$, where the hyperbolic distance $\mathcal{D}_{\mathbb{H}}$ is computed as:
\begin{equation}
\mathcal{D}_{\mathbb{H}}(\mathbf{a},\mathbf{b}) =
\frac{2}{\sqrt{c}} \operatorname{arctanh}\!\left(\sqrt{c}\,\|-\mathbf{a} \oplus_c \mathbf{b}\|\right).
\end{equation}
Unlike angular alignment, hyperbolic distance encodes unique norm-based structural information that is not fully captured by the flat affinity graph. Therefore, to avoid imposing mismatched Euclidean structural priors on the curved geometry, we formulate the distance loss as a standard alignment task supervised by the identity matrix $\mathbb{I}$:
\begin{equation}
\label{eq:dis}
    \mathcal{L}_{\text{dis}}^{\mathbb{H}} = \sum_{v \neq u} \mathcal{L}_{\text{con}}\Big(\mathbf{\hat Q}^{\mathbb{H}}_{v}, \mathbf{Q}^{\mathbb{H}}_{u}; \mathbb{I}; \mathcal{S}_{\text{d}}(\mathbf{\hat Q}^{\mathbb{H}}_{v}, \mathbf{Q}^{\mathbb{H}}_{u})\Big).
\end{equation}
\textit{Geometric Insight:} While angular alignment secures the direction, embeddings may still drift along the {radial dimension}. This distance-based objective acts as a \textit{hierarchical anchor}. By leveraging complete pairs to learn precise geodesic proximities, it ``pins'' instances to their correct hierarchical depths. This not only cultivates a fine-grained hierarchical structure for clustering (distinguishing general vs. specific concepts) but also eliminates the {structural blurriness} in imputed data. It ensures that recovered views are not just directionally correct but also structurally compact, reducing the uncertainty often found in missing modalities.

The final hyperbolic instance-level objective unifies these perspectives into a coherent framework:
\begin{equation}
\label{eq:ins}
    \mathcal{L}_{\text{ins}}^{\mathbb{H}} = \alpha \mathcal{L}_{\text{dis}}^{\mathbb{H}} + (1-\alpha)\mathcal{L}_{\text{ang}}^{\mathbb{H}},
\end{equation}
where $\alpha$ linearly increases with the training epochs. This dynamic weighting strategy allows the model to prioritize angular consistency $\mathcal{L}_{\text{ang}}^{\mathbb{H}}$ to form basic semantic clusters in the early stages, while progressively integrating distance constraints $\mathcal{L}_{\text{dis}}^{\mathbb{H}}$ to refine the hierarchical structure.

\subsubsection{Hyperbolic Prototype-level Semantic Consistency}

While instance-level alignment refines sample-to-sample features, robust IMVC also requires consistent global semantic structures across views. However, standard prototype-based contrastive learning in Euclidean space often leads to blurred boundaries between semantically proximal categories. To overcome this, we introduce a hyperbolic 
prototype-oriented module designed to enforce cross-view alignment of hierarchy-aware prototype distributions.

Leveraging the view-specific hyperbolic prototype heads defined in Eq.~\eqref{eq:hyp_mapping}, we obtain the hierarchy-aware distributions $\mathbf{\hat{P}}^{\mathbb{H}}_v$ and $\mathbf{P}^{\mathbb{H}}_v \in \mathbb{D}_c^{N \times K}$, where each entry represents the soft assignment of an instance to one of the $K$ hyperbolic prototypes. Crucially, $K$ governs the semantic granularity and is not strictly bound to the cluster number. Optimal selection is vital: an insufficient $K$ limits hierarchical depth in complex data, whereas an excessive $K$ induces redundancy in simpler tasks. This confirms that our enhancement sharpens discriminative boundaries when the geometric configuration aligns with the true data structure (see Sec.~\ref{sec:numberofP} for details).

Unlike Euclidean approaches, these distributions inherently encode hierarchical relations among prototype centers. To establish reliable semantic anchors across heterogeneous modalities, we minimize the cross-view discrepancy by maximizing the angular similarity ($S_a$) between corresponding prototype distributions while pushing apart distinct ones:
\begin{equation}
\label{eq:pro}
\mathcal{L}_{\mathrm{pro}}^{\mathbb{H}}\!\! =\!\! - \frac{1}{K}\! \sum_{v \neq u}\! \sum_{k=1}^{K} \!\\\log\! \frac{\exp(\mathcal{S}_{a}\!(\mathbf{\hat{P}}^{\mathbb{H}}_{v,k},\! \mathbf{P}^{\mathbb{H}}_{u,k})\! /\! \tau)}{\sum_{i=1}^{K} \!\exp(\mathcal{S}_{a}\!(\mathbf{\hat{P}}^{\mathbb{H}}_{v,k}, \!\mathbf{P}^{\mathbb{H}}_{u,i})\! /\! \tau)}\!.
\end{equation}
Here, $\mathbf{\hat{P}}^{\mathbb{H}}_{v,k}$ denotes the $k$-th column of the predicted distribution from view $v$, and $\mathbf{P}^{\mathbb{H}}_{u,k}$ is the corresponding column of the teacher distribution from view $u$.
By modeling fine-grained prototype distributions within the non-Euclidean geometry, this module provides high-level semantic guidance for the learned features. This structural alignment not only distinguishes instances according to their intrinsic hierarchy but also regularizes the imputation process, ensuring that missing views are recovered within the correct semantic categories without conflation.

\begin{algorithm}[t]
\small
\caption{Training and Inference of HERL}
\label{alg:HERL_compact}
\begin{algorithmic}[1]
\REQUIRE Incomplete dataset $\mathbf{X}$, hyperparameters.
\ENSURE Clustering assignments.
\WHILE{not converged}
    \STATE \textbf{Euclidean Step:} Extract features $\mathbf{Z}^v,\mathbf{F}^v, \mathbf{F}_t^v$; construct graph $\mathbf{G}^v$; compute $\mathcal{L}_{\text{con}}^{\mathbb{E}}$ via Eq.~\eqref{eq:con}.
    \STATE \textbf{Mapping Step:} Project to Poincaré ball via Eq.~\eqref{eq:hyp_mapping}.
    \STATE \textbf{Instance-level Step:} Compute alignment $\mathcal{L}_{\text{ins}}^{\mathbb{H}}$ combining angular and distance constraints (Eqs.~\ref{eq:ang}-\ref{eq:ins}).
    \STATE \textbf{Prototype-level Step:} Compute hierarchy-aware distributions consistency $\mathcal{L}_{\text{pro}}^{\mathbb{H}}$ via Eq.~\eqref{eq:pro}.
    \STATE \textbf{Update:} Minimize $\mathcal{L}_{\text{total}} = \mathcal{L}_{\text{con}}^{\mathbb{E}} + \mathcal{L}_{\text{ins}}^{\mathbb{H}} + \beta\mathcal{L}_{\mathrm{pro}}^{\mathbb{H}}$; update teacher $\theta_t$ via EMA.
\ENDWHILE
\STATE \textbf{Inference:} Recover missing views via Eq.~\eqref{eq:completion}; perform $k$-means on concatenated features.
\end{algorithmic}
\end{algorithm}

\subsection{Overall Objective}
By integrating the aforementioned loss terms into a unified optimization framework, the overall objective function of the proposed HERL is formulated as follows:
\begin{equation}
\label{eq:overall}
\mathcal{L}
= \mathcal{L}_{\text{con}}^{\mathbb{E}} + \mathcal{L}_{\text{ins}}^{\mathbb{H}} + \beta\mathcal{L}_{\mathrm{pro}}^{\mathbb{H}},
\end{equation}
where $\beta$ denotes a trade-off hyperparameter.

\subsection{Theoretical Justification on Representation Capacity}
\label{sec:theory}

In this section, we provide a rigorous theoretical justification for the geometric advantage of the proposed HERL framework. We analyze the representation capacity mismatch between Euclidean and Hyperbolic spaces when modeling data with latent hierarchical structures (e.g., scene hierarchies in Scene-15, spatial categories in LandUse-21). To rigorously justify this, we model the latent structure of the dataset as a discrete regular tree $\mathcal{T}$ with branching factor $b \ge 2$ and depth $R$.

\noindent\textbf{Preliminaries: Metric Embeddings and Distortion.}
Let $\mathcal{T} = (V, E)$ be the regular tree, where $V$ denotes the set of nodes. The tree metric $d_{\mathcal{T}}(u, v)$ is the shortest path distance between nodes $u, v \in V$. An embedding $f: V \to (\mathcal{M}, d_{\mathcal{M}})$ maps the tree nodes into a metric manifold $\mathcal{M}$. The distortion $D$ of this embedding measures the worst-case stretching or compressing of distances relative to a global scaling factor $s$:
\begin{equation}
    s \cdot d_{\mathcal{T}}(u, v) \le d_{\mathcal{M}}(f(u), f(v)) \le D \cdot s \cdot d_{\mathcal{T}}(u, v), \quad \forall u, v \in V
\end{equation}
If $D=1$, the embedding is isometric (perfectly structure-preserving). A large $D$ implies that semantically distinct nodes are forced to be arbitrarily close, leading to semantic blurring.

\noindent\textbf{Proposition 1 (Geometric Capacity Mismatch).} \textit{Embedding $\mathcal{T}$ into Euclidean space $\mathbb{R}^n$ incurs a distortion $D_{\mathbb{E}}$ that grows exponentially with depth, whereas embedding into the Poincaré ball $\mathbb{D}^2$ maintains a constant distortion $D_{\mathbb{H}}$:}
\begin{equation}
    D_{\mathbb{E}} = \Omega(b^{R/n} \cdot R^{-1}) \quad \text{vs.} \quad D_{\mathbb{H}} \approx 1 + \epsilon.
\end{equation}

\noindent\textit{Proof.} The fundamental disparity stems from the packing bottleneck in Euclidean space versus the natural exponential expansion of hyperbolic space ($\sinh r \sim e^r$). Here, we provide the detailed proof for this capacity mismatch, divided into the impossibility result for Euclidean space and the existence result for Hyperbolic space.

\noindent\textbf{Part I: The Euclidean Lower Bound (Impossibility Result).} \textit{Statement:} Any embedding of a regular tree $\mathcal{T}$ (branching factor $b$, depth $R$) into a Euclidean space $\mathbb{R}^n$ of fixed dimension $n$ incurs a distortion $D_{\mathbb{E}}$ that grows exponentially with depth $R$: $D_{\mathbb{E}} = \Omega(b^{R/n} \cdot R^{-1})$.

\noindent\textit{Intuition:} The number of nodes in a tree grows exponentially ($b^R$), while the volume of a Euclidean ball grows only polynomially with its radius ($r^n$). To pack exponentially many nodes into a polynomial container, the nodes must be aggressively compressed together, destroying the distance structure and leading to unavoidable semantic blurring.

We employ a volumetric packing argument focusing on the leaf nodes of the tree.
\begin{enumerate}
    \item \textbf{The Demand (Exponential Growth):} The number of leaves in $\mathcal{T}$ is $|L| = b^R$. In the tree metric, any two distinct leaves $u, v$ are separated by at least $d_{\mathcal{T}}(u, v) \ge 2$. For an embedding to preserve distinctness with scaling factor $s$, we must be able to place disjoint Euclidean balls of radius $s$ centered at each leaf image $f(u)$.
    \begin{equation}
        \text{Required Volume} = |L| \cdot \text{Vol}(\mathcal{B}_s^n) = b^R \cdot C_n s^n
    \end{equation}
    where $C_n$ is the volume coefficient for the $n$-dimensional unit ball.
    \item \textbf{The Supply (Polynomial Constraint):} The maximum distance in the tree is $2R$ (diameter). With distortion $D$, the maximum distance in the embedding is bounded by $2D \cdot s \cdot R$. Thus, the entire set of embedded points must fit within a Euclidean ball of radius $R_{max} \approx D \cdot s \cdot R$.
    \begin{equation}
        \text{Available Volume} = \text{Vol}(\mathcal{B}_{R_{max}}^n) = C_n (D \cdot s \cdot R)^n
    \end{equation}
    \item \textbf{The Contradiction:} Packing condition requires $\text{Total Required Volume} \le \text{Available Volume}$:
    \begin{equation}
        b^R \cdot C_n s^n \le C_n (D \cdot s \cdot R)^n
    \end{equation}
    Dividing by $C_n s^n$ and taking the $n$-th root:
    \begin{equation}
        b^{R/n} \le D \cdot R \implies D \ge \frac{b^{R/n}}{R}
    \end{equation}
\end{enumerate}
\textbf{Conclusion:} As the hierarchy depth $R$ increases, the distortion $D$ explodes exponentially. This theoretical lower bound explains why Euclidean-based baselines suffer from mode collapse on complex datasets: the space simply lacks the capacity to separate the hierarchical clusters.

\noindent\textbf{Part II: The Hyperbolic Upper Bound (Existence Result).} \textit{Statement:} For any regular tree $\mathcal{T}$, there exists an embedding $f: \mathcal{T} \to \mathbb{D}^2$ (2D Poincaré disk) such that the distortion $D_{\mathbb{H}}$ is bounded by a small constant ($D_{\mathbb{H}} \approx 1+\epsilon$), independent of depth $R$.

\noindent\textit{Intuition:} In hyperbolic space, the circumference of a circle grows exponentially with its radius ($L \propto e^r$). This expansion perfectly matches the exponential growth of tree nodes, allowing us to place nodes recursively without running out of room. We use the constructive proof by~\cite{DBLP:conf/gd/Sarkar11} in the Poincaré disk with curvature $-c=1$.

\begin{enumerate}
    \item \textbf{Construction Strategy (Recursive Placement):} We place the root at the origin. We place children of a node $u$ (at radius $r_k$) on a circle at radius $r_{k+1} = r_k + \tau$, where $\tau$ is the radial step size. The angular sector available for a node at level $k$ shrinks exponentially as the tree branches:
    \begin{equation}
        \Delta \theta_k \approx \frac{2\pi}{b^k}
    \end{equation}
    In Euclidean space, this shrinking angle would force nodes to become arbitrarily close. In Hyperbolic space, we can compensate for this by increasing the radius.
    \item \textbf{The Separation Condition:} To maintain isometry, the hyperbolic distance $d_{\mathbb{H}}$ between adjacent siblings at level $k+1$ must be maintained at $\approx \tau$ (the edge length). Using the \textit{Hyperbolic Law of Cosines} for a triangle with two sides $r_{k+1}$ and angle $\Delta \theta_{k+1}$:
    \begin{equation}
        \cosh(d_{\mathbb{H}}) = \cosh^2(r_{k+1}) - \sinh^2(r_{k+1})\cos(\Delta \theta_{k+1})
    \end{equation}
    Using the approximation $\cosh(d) \approx \frac{1}{2}e^d$, $\sinh(r) \approx \frac{1}{2}e^r$, and $\cos(\theta) \approx 1-\theta^2/2$ for large $r$ and small $\theta$, this simplifies to:
    \begin{equation}
    \begin{split}
        d_{\mathbb{H}} &\approx 2 \ln \left( \sinh(r_{k+1}) \cdot \frac{\Delta \theta_{k+1}}{2} \right) \\
        &\approx 2 \ln \left( \frac{1}{2}e^{r_{k+1}} \cdot \frac{\pi}{b^{k+1}} \right)
    \end{split}
    \end{equation}
    \item \textbf{The Exponential Cancellation:} We set the radius to grow linearly: $r_{k+1} = (k+1)\tau$. Substituting this into the equation:
    \begin{equation}
        d_{\mathbb{H}} \approx 2 \ln \left( C \cdot \frac{e^{(k+1)\tau}}{b^{k+1}} \right) = 2 \ln \left( C \cdot \left( \frac{e^\tau}{b} \right)^{k+1} \right)
    \end{equation}
    To prevent the distance from collapsing (i.e., to keep $d_{\mathbb{H}}$ constant), we choose the step size $\tau$ such that the base of the exponential term is 1:
    \begin{equation}
        \frac{e^\tau}{b} = 1 \implies \tau = \ln b
    \end{equation}
\end{enumerate}
\textbf{Conclusion:} By setting the radial step $\tau = \ln b$, the exponential expansion of the hyperbolic space ($e^\tau$) perfectly cancels the exponential reduction in angular space ($b^{-1}$). Consequently, the pairwise distances between siblings remain constant regardless of the hierarchy depth $k$. This proves that HERL can disentangle fine-grained categories at arbitrary depths without semantic drift.

\noindent\textbf{Implication.} This geometric alignment empowers HERL to reconstruct latent hierarchical structures even from limited complete views, mitigating the semantic ambiguity caused by data incompleteness. By providing reliable geometric anchors that resist collapse, the framework yields discriminative representations with sharp boundaries and guides robust cross-view completion, ensuring that recovered data adheres to the underlying taxonomy without overcrowding.

\section{Experiments}

\begin{table*}[!ht]
\centering
    \caption{The clustering performance on four multi-view benchmarks with different missing rates. The best and the second-best results are denoted in \textbf{bold} and {\ul underline}, respectively.}
    \label{table:main}
\setlength{\tabcolsep}{2.3pt}
\small
\begin{tabular}{|c|c|ccc|ccc|ccc|ccc|ccc|}
\hline
\multicolumn{1}{|l|}{\multirow{2}{*}{Datasets}}                 & Missing rates & \multicolumn{3}{c|}{0.0}                         & \multicolumn{3}{c|}{0.1}                         & \multicolumn{3}{c|}{0.3}                         & \multicolumn{3}{c|}{0.5}                         & \multicolumn{3}{c|}{0.7}                         \\ \cline{2-17} 
\multicolumn{1}{|l|}{}                                          & Metrics       & ACC            & NMI            & ARI            & ACC            & NMI            & ARI            & ACC            & NMI            & ARI            & ACC            & NMI            & ARI            & ACC            & NMI            & ARI            \\ \hline
\multirow{9}{*}{\rotatebox{90}{Scene-15}}                            & COMPLETER               & 39.28          & 42.97          & 22.85          & 41.69          & 43.85          & 24.90          & 40.00          & 43.03          & 23.82          & 39.16          & 41.96          & 22.90          & 37.62          & 39.96          & 22.20          \\

                                                     & SURE                    & 43.03          & 42.65          & 25.15          & 41.85          & 44.48          & 25.72          & 38.46          & 41.33          & 23.47          & 37.90          & 40.13          & 21.82          & 43.03          & 43.44          & 25.91          \\
                                                     & DSIMVC        & 29.74          & 35.44          & 17.52          & 30.62          & 35.72          & 17.04          & 31.57          & 36.49          & 17.75          & 30.72          & 35.64          & 17.14          & 29.58          & 34.68          & 16.39          \\                                        
                                                     & DIMVC                   & 37.28          & 37.49          & 19.73          & 34.41          & 36.43          & 17.84          & 34.05          & 34.97          & 16.92          & 30.07          & 31.86          & 13.96          & 29.23          & 29.87          & 13.63          \\
                                                     & ProImp                  & 43.60          & 45.00          & 26.80          & 43.56          & 44.57          & 26.70          & 40.77          & 43.20          & 25.05          & 41.60          & 42.90          & 25.30          & 40.89          & 41.45          & 24.06          \\
                                                     & ICMVC                   & 39.78          & 37.42          & 22.48          & 39.21          & 36.81          & 21.99          & 37.40          & 35.00          & 20.58          & 31.39          & 27.80          & 15.02          & 25.79          & 24.52          & 11.70          \\
                                                     & DCG                     & 36.21          & 35.43          & 19.40          & 37.53          & 34.65          & 20.16          & 38.08          & 35.57          & 21.51          & 32.84          & 30.68          & 17.64          & 33.60          & 29.92          & 17.35          \\
                                                     & DIVIDE                  & {\ul 47.51}    & {\ul 48.63}    & {\ul 30.97}    & {\ul 47.40}    & {\ul 47.62}    & {\ul 30.53}    & {\ul 47.02}    & {\ul 47.00}    & {\ul 30.08}    & {\ul 45.42}    & {\ul 45.51}    & {\ul 29.09}    & {\ul 43.34}    & {\ul 43.21}    & {\ul 26.79}    \\
                                                     & \textbf{HERL}                    & \textbf{50.33} & \textbf{49.22} & \textbf{32.97} & \textbf{49.86} & \textbf{49.18} & \textbf{32.66} & \textbf{49.94} & \textbf{48.48} & \textbf{32.36} & \textbf{48.75} & \textbf{47.14} & \textbf{31.26} & \textbf{45.96} & \textbf{45.20} & \textbf{28.84} \\ \hline
\multirow{9}{*}{\rotatebox{90}{Reuters}}                             & COMPLETER               & 39.75          & 22.98          & 9.82           & 40.91          & 24.54          & 8.26           & 40.77          & 23.46          & 9.41           & 39.14          & 23.75          & 10.95          & 40.99          & 21.88          & 8.88           \\
                                                     & SURE                    & 49.35          & 28.94          & 23.01          & 46.55          & 28.61          & 24.85          & 43.73          & 27.96          & 20.53          & 51.75          & 30.90          & 26.04          & 47.82          & 33.12          & 27.84          \\
                                                     & DSIMVC        & 45.44          & 28.07          & 23.71          & 43.72          & 24.62          & 20.71          & 42.42               & 22.69               & 19.24               & 42.02               & 21.22               & 18.58               & 37.47               & 17.36               & 14.73               \\
                                                     & DIMVC                   & 51.28          & 26.53          & 20.55          & 48.88          & 22.79          & 17.63          & {\ul 52.28}    & 27.23          & 21.21          & 53.62          & 26.25          & 20.87          & 49.44          & 22.31          & 18.58          \\
                                                     & ProImp                  & {\ul 56.54}          & {\ul 39.35}          & {\ul 32.77}    & {\ul 53.25}    & {\ul 39.10}         & {\ul 32.07}    & 47.21          & {\ul 35.26}          & {\ul 27.00}    & 51.89          & 35.54          & {\ul 28.53}          & 47.40          & 34.64          & 26.58          \\
                                                     & ICMVC                   & 48.24          & 33.70          & 25.57          & 49.07          & 33.57          & 25.59          & 48.79          & 33.94          & 26.09          & 47.59          & 28.43          & 23.56          & 51.91          & 32.29          & 26.15          \\
                                                     & DCG                     & 50.54          & 35.59          & 27.81          & 52.63          & {37.37}    & 29.42          & 49.02          & 17.73          & 19.61          & 48.53          & 33.87          & 27.80          & 46.83          & 30.00          & 24.44          \\
                                                     & DIVIDE                  & {45.08}    & {30.85}    & 18.21          & 50.90          & 34.27          & 22.94          & 47.96          & 32.49    & 20.36          & {\ul 53.68}    & {\ul 37.44}    & {28.14}    & {\ul 53.88}    & {\ul 36.64}    & {\ul 28.10}    \\
                                                     & \textbf{HERL}                    & \textbf{59.12} & \textbf{39.45} & \textbf{33.12} & \textbf{59.82} & \textbf{39.58} & \textbf{34.13} & \textbf{59.12} & \textbf{39.23} & \textbf{33.83} & \textbf{59.14} & \textbf{38.13} & \textbf{33.50} & \textbf{55.87} & \textbf{38.19} & \textbf{32.73} \\ \hline
\multirow{9}{*}{\rotatebox{90}{LandUse-21}}                          & COMPLETER               & 25.06          & 31.98          & 13.34          & 25.57          & 32.12          & 13.49          & 26.01          & 31.15          & 12.35          & 20.59          & 26.25          & 9.92           & 24.88          & 28.81          & 11.61          \\
                                                     & SURE                    & 25.62          & 30.51          & 11.93          & 27.00          & 29.79          & 12.31          & 24.00          & 28.76          & 11.00          & 26.71          & 29.21          & 13.09          & 25.05          & 25.88          & 10.88          \\
                                                     & DSIMVC        & 19.19          & 18.53          & 5.86           & 18.00          & 18.79          & 5.39           & 17.94          & 17.87          & 5.12           & 17.56          & 17.35          & 4.85           & 16.83          & 16.69          & 4.54           \\
                                                     & DIMVC                   & 25.81          & 32.21          & 12.38          & 23.34          & 27.80           & 9.59           & 23.39          & 29.88          & 10.59          & 23.61          & 29.62          & 10.07          & 22.68          & 26.80           & 9.06           \\
                                                     & ProImp                  & 23.70          & 27.90          & 10.80          & 24.43          & 29.22          & 11.44          & 24.63          & 28.45          & 11.78          & 24.45          & 27.43          & 10.94          & 23.86          & 27.05          & 10.82          \\
                                                     & ICMVC                   & 27.02          & 30.86          & 14.12          & 26.36          & 30.52          & 13.40          & 26.23          & 29.64          & 13.26          & 24.51          & 26.40          & 10.71          & 21.47          & 24.33          & 8.70          \\
                                                     & DCG                     & 26.43          & 30.31          & 13.43          & 26.81          & 30.71          & 13.60          & 26.12          & 29.83          & 12.87          & 23.52          & 26.94          & 10.45          & 22.97          & 25.15          & 9.65           \\
                                                     & DIVIDE                  & {\ul 32.45}    & {\ul 39.77}    & {\ul 17.97}    & {\ul 30.51}    & {\ul 35.97} & {\ul 16.10}    & {\ul 30.40}    & {\ul 35.77}    & {\ul 16.17}    & {\ul 29.86}    & {\ul 35.23}    & {\ul 15.73}    & {\ul 29.88}    & {\ul 34.74}    & {\ul 15.55}    \\
                                                     & \textbf{HERL}                    & \textbf{33.12} & \textbf{40.94} & \textbf{18.87} & \textbf{32.23} & \textbf {37.57}    & \textbf{17.16} & \textbf{31.74} & \textbf{36.81} & \textbf{16.84} & \textbf{31.20} & \textbf{36.85} & \textbf{16.93} & \textbf{30.98} & \textbf{34.99} & \textbf{16.13} \\ \hline
\multirow{9}{*}{\rotatebox{90}{HandWritten}}                         & COMPLETER               & 76.75          & 79.87          & 65.32          & 75.36          & 79.86          & 68.26          & 77.97          & 81.14          & 70.40          & 74.54          & 77.37          & 64.30          & 74.52          & 77.48          & 67.44          \\
                                                     & SURE                    & 89.35          & 79.80          & 78.22          & 85.40          & 75.80          & 71.35          & 69.85          & 61.58          & 51.91          & 75.75          & 62.37          & 55.44          & 67.25          & 59.83          & 50.43          \\
                                                     & DSIMVC        & 82.80          & 79.25          & 73.15          & 78.40          & 75.26          & 67.92          & 73.29          & 70.37          & 60.96          & 66.39          & 63.49          & 51.21          & 53.86          & 51.15          & 36.92          \\
                                                     & DIMVC                   & 66.75          & 63.29          & 52.86          & 77.15          & 69.18          & 61.71          & 59.76          & 46.81          & 36.40          & 79.24          & 67.97          & 61.36          & 56.78          & 54.13          & 42.22          \\
                                                     & ProImp                  & 83.23          & 81.42          & 74.55          & 82.45          & 80.58          & 73.40          & 83.04          & 80.49          & 74.48          & 78.98          & 74.08          & 66.47          & 78.84          & 73.65          & 66.68          \\
                                                     & ICMVC                   & 87.52          & 78.40          & 75.11          & 87.71          & 79.07          & 75.63          & 84.10          & 74.53          & 70.20          & 77.24          & 68.32          & 60.16          & 58.19          & 50.17          & 36.60          \\
                                                     & DCG                     & 85.65          & 85.41          & 78.96          & 82.75          & 82.63          & 74.88          & 82.70          & 80.54          & 73.96          & 80.80          & 76.21          & 70.45          & 79.75          & 74.00          & 67.54          \\
                                                     & DIVIDE                  & {\ul 92.78}    & {\ul 86.65}    & {\ul 84.93}    & {\ul 89.58}    & {\ul 85.14}    & {\ul 82.13}    & {\ul 87.80}    & {\ul 82.34}    & {\ul 78.36}     & {\ul 86.53}    & {\ul 77.23}    & {\ul 73.71}    & {\ul 84.41}    & {\ul 73.82}    & {\ul 69.65 }    \\
                                                     & \textbf{HERL}                    & \textbf{94.94} & \textbf{89.53} & \textbf{89.13} & \textbf{93.65} & \textbf{87.36} & \textbf{86.49} & \textbf{92.49} & \textbf{85.80} & \textbf{84.36} & \textbf{89.89} & \textbf{81.74} & \textbf{79.54} & \textbf{86.82} & \textbf{77.29} & \textbf{74.01} \\ \hline
\end{tabular}

\end{table*}

\begin{figure*}[!ht]
    \centering 

    \begin{subfigure}[c]{0.28\textwidth} 
        \centering
        \includegraphics[width=\linewidth]{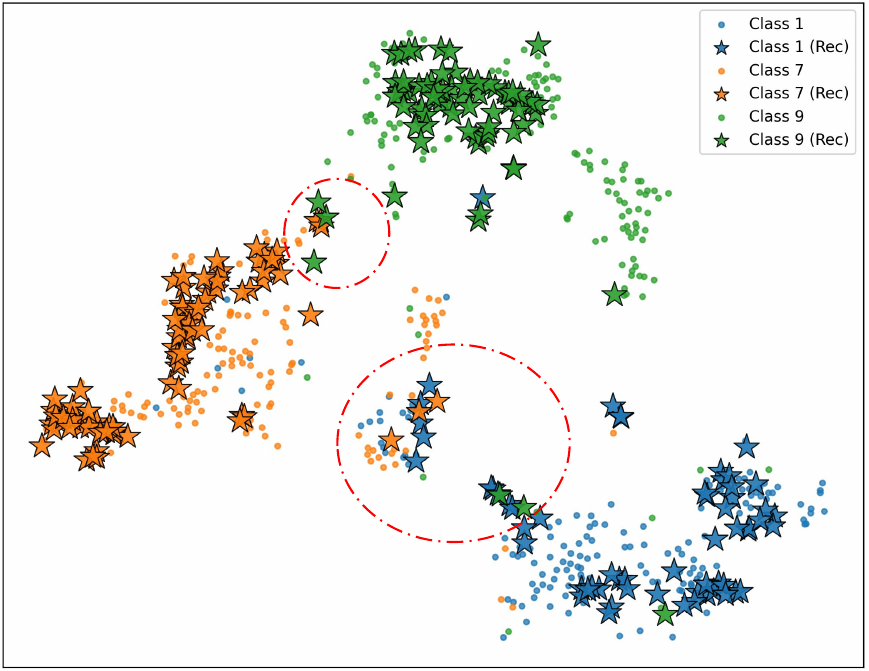}\\
        \includegraphics[width=\linewidth]{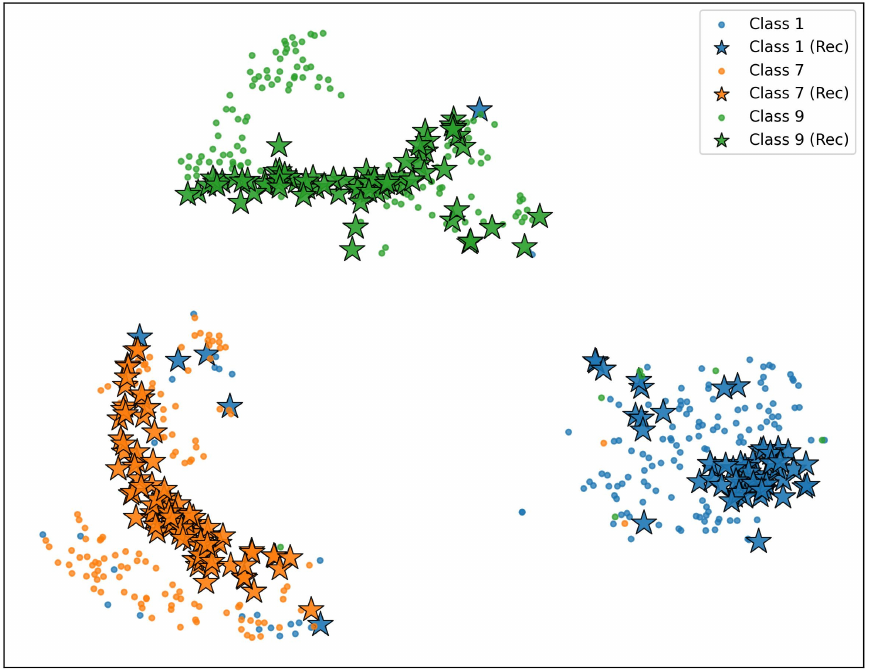}
        \caption{Imputation results on View 1.}
        \label{fig:4a}
    \end{subfigure}
    \hspace{0.25cm} 
    \begin{subfigure}[c]{0.57\textwidth}
        \centering
        \includegraphics[width=\linewidth]{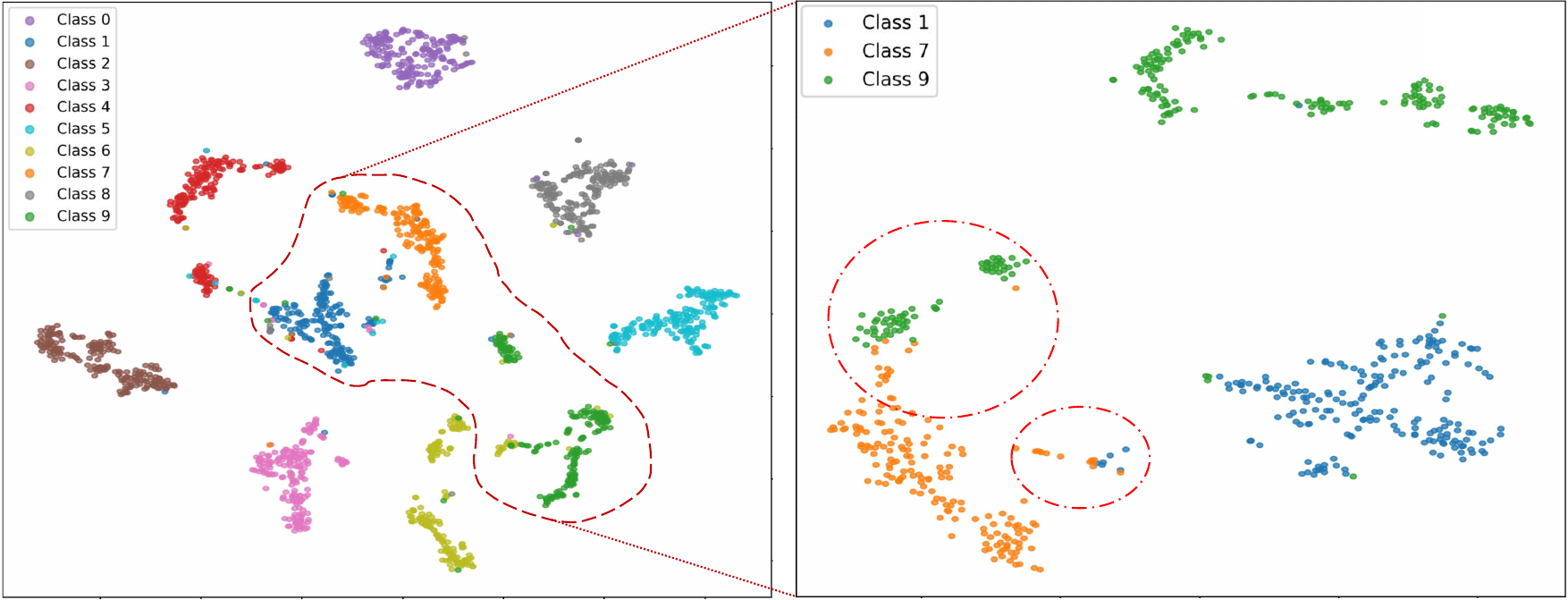}\\
        \includegraphics[width=\linewidth]{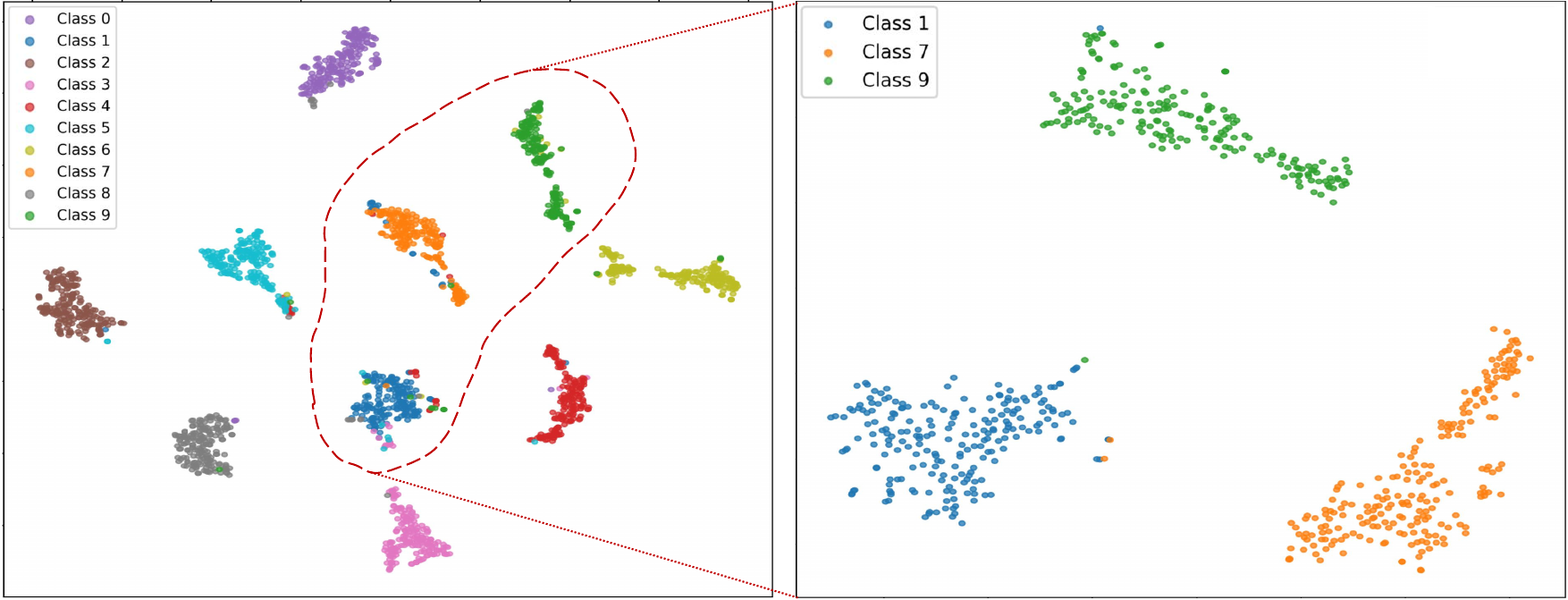}
        \caption{Global and local t-SNE visualizations.}
        \label{fig:4b}
    \end{subfigure}

    \caption{{Qualitative analysis on the HandWritten dataset.} The top and bottom rows represent Euclidean-based method DIVIDE and our HERL, respectively. (a) Comparison of imputation performance on View 1 under severe incompleteness ($\eta=0.7$) to highlight recovery robustness. (b) Global t-SNE visualization (left) and local structural alignment (right) for similar digits 1, 7, and 9 at $\eta=0.1$.}
\label{fig:tsne}
\end{figure*}

\subsection{Experimental Setup}\label{sec:implementation}



\noindent\textbf{Datasets and Evaluation Metrics.}
We conduct extensive experiments on four widely-used benchmark datasets. As mentioned in the main text, we select datasets that exhibit inherent hierarchical structures to fully leverage the advantages of hyperbolic geometry.

\subsubsection{\textbf{Scene-15} \cite{Scene15}} This dataset consists of 4,485 images distributed across 15 scene categories, where \textit{GIST} and \textit{LBP} features serve as two views. The categories follow a semantic taxonomy, ranging from broad classifications (e.g., indoor vs. outdoor) to fine-grained scenes (e.g., bedroom, kitchen, forest, mountain), which naturally form a hierarchical structure.

\subsubsection{\textbf{Reuters} \cite{Reuters}} This is a multilingual news dataset containing 18,758 samples. We utilize 10-dimensional English and French textual latent embeddings as two views. The textual data possesses an intrinsic topical hierarchy, where general topics branch into more specific sub-events.

\subsubsection{\textbf{LandUse-21} \cite{LandUse}} This dataset comprises 2,100 satellite images from 21 classes, with \textit{PHOG} and \textit{LBP} features as two views. Its hierarchy arises from geospatial semantics, organizing from high-level terrain types to more specific land-use categories.

\subsubsection{\textbf{HandWritten} \cite{HandWritten}} This dataset contains 2,000 digit images (0--9). We construct a two-view setting using the \textit{Fourier coefficients} and the \textit{Karhunen--Loeve coefficients}. Beyond class labels, the dataset also exhibits morphological hierarchies, where digits sharing similar structural patterns form latent tree-like relations in the feature space.




We report standard clustering evaluation metrics, including accuracy (ACC), normalized mutual information (NMI), and adjusted Rand index (ARI), to assess clustering quality.



\noindent\textbf{Baseline Methods.}
We compare HERL with eight state-of-the-art baselines. These methods cover both standard multi-view clustering and incomplete multi-view clustering settings.
\setcounter{subsubsection}{0}
\subsubsection{\textbf{COMPLETER} \cite{COMPLETER}} 
COMPLETER unifies consistent representation learning and cross-view data recovery within an information-theoretic framework. It learns informative shared representations by maximizing mutual information across views and recovers missing views through dual prediction based on conditional entropy minimization.
\subsubsection{\textbf{SURE} \cite{SURE}}
SURE addresses both partially view-unaligned and partially sample-missing problems under a unified contrastive learning framework. It further introduces a noise-robust contrastive loss to alleviate the false negative issue caused by random cross-view negative sampling.
\subsubsection{\textbf{DSIMVC} \cite{DSIMVC}}
DSIMVC proposes a safe incomplete multi-view clustering framework that dynamically imputes missing views from learned semantic neighbors. Through a bi-level optimization strategy, it also automatically selects reliable imputed samples for training to reduce the risk of performance degradation caused by semantically inconsistent imputations.
\subsubsection{\textbf{DIMVC} \cite{DIMVC}}
DIMVC is an imputation-free and fusion-free deep IMVC method that learns view-specific representations and clustering structures separately for each view. It exploits multi-view cluster complementarity in a high-dimensional space and adopts an EM-like optimization scheme to iteratively refine feature learning and clustering.
\subsubsection{\textbf{ProImp} \cite{ProImp}}
ProImp introduces a dual-stream framework to capture both instance commonality and view versatility in incomplete multi-view clustering. By learning view-specific prototypes and sample--prototype relationships via dual attention and dual contrastive learning, it enables missing-view recovery based on prototype information.
\subsubsection{\textbf{ICMVC} \cite{ICMVC}}
ICMVC is an end-to-end incomplete contrastive multi-view clustering framework with high-confidence guidance. It combines consistency relation transfer, graph convolution, attention-based fusion, and instance-level contrastive learning to jointly handle missing views, representation learning, and clustering.
\subsubsection{\textbf{DIVIDE} \cite{DIVIDE}}
DIVIDE is a robust contrastive multi-view clustering method that employs high-order random walks to identify positive and negative pairs in a more global manner. It further decouples inter-view and intra-view contrastive learning into different embedding spaces, improving robustness against false pairs and missing views.
\subsubsection{\textbf{DCG} \cite{DCG}}
DCG is a diffusion-based incomplete multi-view clustering method that performs view recovery through forward diffusion and reverse denoising processes. It further combines instance-level and category-level interactive learning to exploit both consistent and complementary information for robust end-to-end clustering.

\noindent\textbf{Implementation Details.}
HERL is implemented in PyTorch on RTX 4090 GPUs using Adam~\cite{adam}.
We set the default batch size to 1024 and learning rate to $2 \times 10^{-3}$, adjusting them to 256 and $5 \times 10^{-4}$ for HandWritten.
Training runs for 500 epochs (200 for Reuters), with the relational affinity graph introduced after epoch 100.
Other hyperparameters $(\alpha, \beta, c, cr, K)$ are detailed in the supplemental material.
To simulate incomplete multi-view data, we randomly remove one view from instances based on a missing rate $\eta=m/n$, where $m$ is the number of incomplete instances and $n$ is the total number of samples.

\begin{table}[!t]
    \centering
    \small
    \caption{Ablation study on Reuters with $\eta=0.3$.}
    \label{tab:ablation}
    
    \begin{tabular}{cccc|ccc}
\toprule
\multirow{2}{*}{$\mathcal{L}_{\text{con}}^{\mathbb{E}}$} &
  \multirow{2}{*}{$\mathcal{L}_{\text{ang}}^{\mathbb{H}}$} &
  \multirow{2}{*}{$\mathcal{L}_{\text{dis}}^{\mathbb{H}}$} &
  \multirow{2}{*}{$\mathcal{L}_{\text{pro}}^{\mathbb{H}}$} &
  \multicolumn{3}{c}{Reuters} \\
             &              &                      &              & ACC            & NMI            & ARI            \\ \midrule
$\checkmark$ &              &                      &              & 51.06          & 34.71          & 24.05          \\
$\checkmark$ & $\checkmark$ & \multicolumn{1}{l}{} &              & 52.76          & 35.12          & 25.61          \\
$\checkmark$ &              & $\checkmark$         &              & 51.01          & 35.98          & 25.33          \\
$\checkmark$ & $\checkmark$ & $\checkmark$         &              & 54.24          & 38.04          & 27.83          \\
$\checkmark$ &              &                      & $\checkmark$ & 56.76          & 37.20          & 32.22          \\
$\checkmark$ & $\checkmark$ & $\checkmark$         & $\checkmark$ & \textbf{59.12} & \textbf{39.23} & \textbf{33.83} \\ \bottomrule
\end{tabular}

\end{table}

\begin{figure*}[!ht]
    \centering 

    \begin{subfigure}[c]{0.46\textwidth} 
        \centering
        \includegraphics[width=\linewidth]{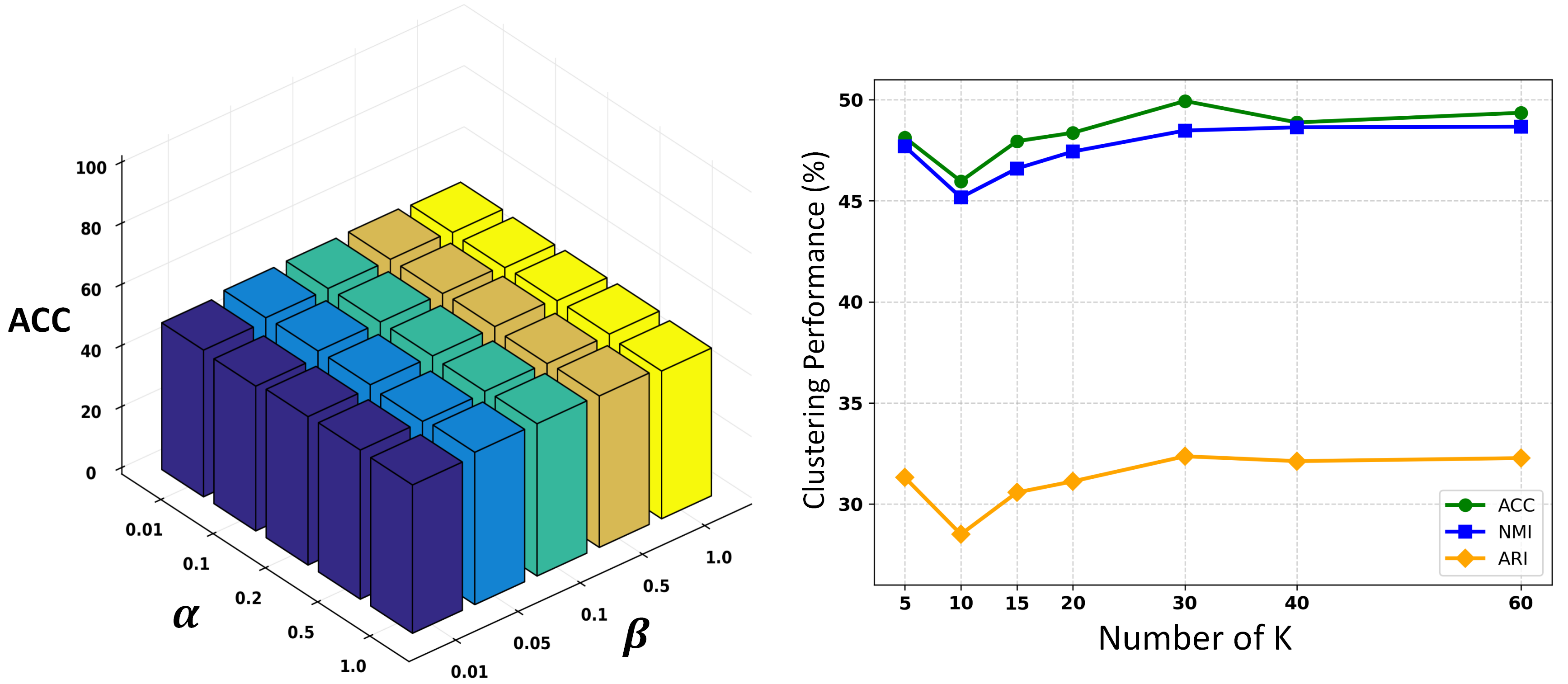}
        \caption{Scene-15}
    \end{subfigure}
    \hspace{0.1cm} 
    \begin{subfigure}[c]{0.46\textwidth}
        \centering
        \includegraphics[width=\linewidth]{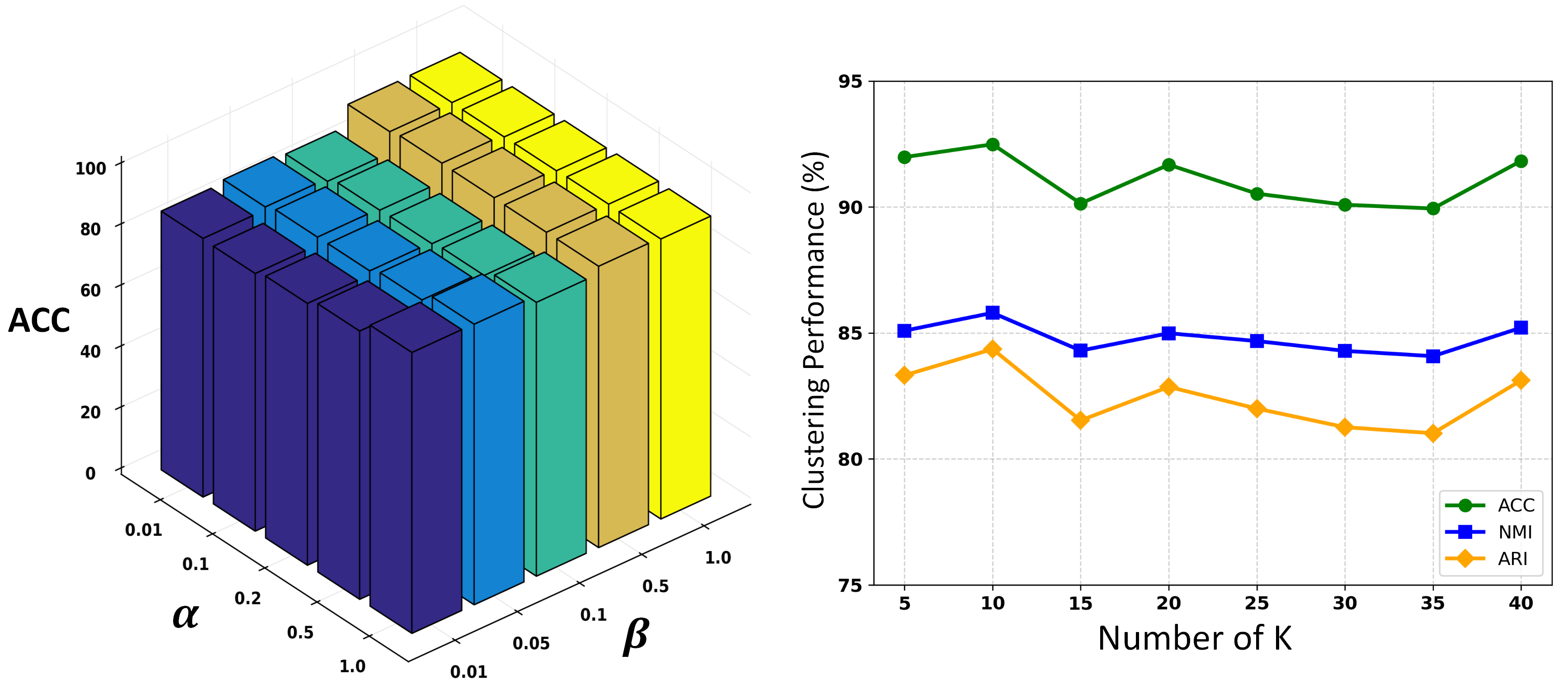}
        \caption{HandWritten}
    \end{subfigure}

    \caption{{Hyperparameter sensitivity analysis on Scene-15 (a) and HandWritten (b) with $\eta=0.3$.} For each dataset, the {left panel} visualizes the joint impact of $\alpha$ and $\beta$, while the {right panel} illustrates the performance variations with respect to the number of prototypes. }
    \label{fig:hypera}
\end{figure*}
\subsection{Main Empirical Results}
\label{sec:exp_main}

\noindent\textbf{HERL achieves SOTA results.} 
As shown in Table~\ref{table:main}, HERL outperforms all rivals by a substantial margin \textbf{across all datasets}. 
Specifically, with the missing rate set to 0.3, our method achieves remarkable performance gains on both Reuters and HandWritten datasets, improving upon the best baseline by \textbf{6.84\%} and \textbf{4.69\%} in ACC, \textbf{4.21\%} and \textbf{3.46\%} in NMI, and \textbf{6.83\%} and \textbf{6.00\%} in ARI, respectively.
Remarkably, HERL ($\eta=0.3$) achieves results comparable to or even surpassing the complete-view baseline ($\eta=0.0$). For instance, on Scene-15, HERL outperforms the baseline in ACC (\textbf{49.94\%} vs. \textbf{47.51\%}), demonstrating that our hyperbolic strategy not only ensures high-quality geometry-driven recovery but also yields superior representations.

\noindent\textbf{HERL learns more distinguishable representations.} 
To qualitatively validate our {hyperbolic enhanced strategy}, we compare HERL against the Euclidean baseline DIVIDE on the HandWritten dataset (Fig.~\ref{fig:tsne}). Regarding imputation (Fig.~\ref{fig:tsne}(a)), we evaluate the models under a challenging missing rate of 0.7 to emphasize their recovery capabilities. The baseline exhibits blurring artifacts indicative of {semantic drift}, whereas HERL recovers distinct and faithful patterns. This confirms that the {hyperbolic structural constraints} provide reliable guidance for inferring missing modalities. As for clustering performance (Fig.~\ref{fig:tsne}(b), $\eta=0.1$), HERL achieves sharper class separation. Particularly for similar digits 1, 7, and 9, the Euclidean baseline suffers from {feature entanglement}, while HERL disentangles these proximal categories. These results verify that the hyperbolic enhancement strategy offers the {structural distinctiveness} to resolve ambiguity in both data recovery and fine-grained clustering.

\begin{table}[!t]
\centering
\small
\caption{Hyperbolic-aware parameters analysis with $\eta=0.3$.}
\label{tab:sensitivity_hyper}
\resizebox{\linewidth}{!}{
\begin{tabular}{c|c|cc|cc|cc}
\toprule 
\multirow{2}{*}{$c$}  & \multirow{2}{*}{$cr$} & \multicolumn{2}{c|}{Scene-15} & \multicolumn{2}{c}{HandWritten} & \multicolumn{2}{c}{LandUse-21} \\ 
                      &                       & ACC            & NMI            & ACC             & NMI            & ACC             & NMI \\ \midrule
\multirow{3}{*}{0.01} & 1.0                   & 49.69          & 48.34          & 90.82           & 84.81          & 30.88                   & \textbf{37.10}          \\
                      & 1.5                   & \textbf{49.94} & 48.48          & 92.16           & 85.31          & 31.03                   & 36.68                   \\
                      & 2.0                   & 49.91          & 48.72          & \textbf{92.49}  & \textbf{85.80} & 31.20                   & 36.87                   \\ \midrule
\multirow{3}{*}{0.05} & 1.0                   & 49.25          & 48.46          & 90.78           & 84.69          & 30.89                   & 36.74                   \\
                      & 1.5                   & 49.73          & 48.22          & 92.05           & 85.25          & 30.99                   & 36.98                   \\
                      & 2.0                   & 49.83          & \textbf{48.75} & 92.40           & 85.70          & 31.30                   & 36.88                   \\ \midrule
\multirow{3}{*}{0.1}  & 1.0                   & 49.37          & 48.40          & 90.74           & 84.64          & \textbf{31.74}          & 36.81                   \\
                      & 1.5                   & 49.37          & 48.40          & 92.17           & 85.28          & 31.33                   & 36.85                   \\
                      & 2.0                   & 49.57          & 48.34          & 91.97           & 85.09          & 30.93                   & 36.99                   \\ \bottomrule
\end{tabular}
}

\end{table}

\subsection{Ablation Studies}
\label{sec:ablation}

Tab.~\ref{tab:ablation} details ablation results on Reuters ($\eta=0.3$) using the Euclidean loss $\mathcal{L}_{\text{con}}^{\mathbb{E}}$ as the baseline. We applied identical dynamic weighting ($\alpha$) across all variants.

\noindent\textbf{Synergy of Angular and Distance Constraints.}
Deploying $\mathcal{L}_{\text{ang}}^{\mathbb{H}}$ or $\mathcal{L}_{\text{dis}}^{\mathbb{H}}$ in isolation yields suboptimal results, occasionally degrading performance. This underscores the {intrinsic geometric coupling} of the Poincaré ball: angular alignment establishes directional semantic identity but lacks hierarchical regularization, while distance anchoring captures depth but may struggle with semantic separation without directional guidance. However, their combination (Row 4) achieves significant improvements, confirming that these objectives are {mutually reinforcing}—jointly ensuring accurate semantic positioning within the curved geometry.

\noindent\textbf{Impact of Hyperbolic Prototype Consistency.}
Incorporating $\mathcal{L}_{\text{pro}}^{\mathbb{H}}$ delivers the most substantial individual gains (e.g., \textbf{+5.70\%} ACC on Reuters). This indicates that instance-level refinement alone is insufficient. By aligning hierarchy-aware prototype distributions, $\mathcal{L}_{\text{pro}}^{\mathbb{H}}$ {sharpens global cluster boundaries} and reduces semantic ambiguity, acting as a critical regularizer against representation drift. Ultimately, the full HERL framework achieves superior performance, validating the necessity of unifying fine-grained geometric constraints with global structural consistency.

\subsection{Hyperparameter Sensitivity Analysis}

We investigate the sensitivity of HERL to key hyperparameters in Fig.~\ref{fig:hypera} and Tab.~\ref{tab:sensitivity_hyper} 

\noindent \textbf{Loss Coefficients ($\alpha$ and $\beta$).} 
These coefficients play a pivotal role in balancing the multi-objective optimization framework. 
Specifically, $\alpha$ regulates the trade-off within the instance-level module between \textit{Angular Alignment} (semantic direction) and \textit{Distance-based Anchoring} (hierarchical depth). 
Simultaneously, $\beta$ controls the contribution of the global \textit{Prototype-level Semantic Consistency}.
As illustrated in Fig.~\ref{fig:hypera}, HERL achieves stable and superior performance across a broad spectrum of values. 
However, extreme settings lead to performance degradation. 
The experimental results suggest that HERL is robust to these weights provided they are within a reasonable magnitude, avoiding the need for meticulous fine-tuning.

\noindent \textbf{Hyperbolic-aware Parameters ($c$ and $cr$).}
We analyze the sensitivity of the manifold curvature $c$ and the boundary clipping threshold $cr$ in Tab.~\ref{tab:sensitivity_hyper}. 
Curvature $c$ governs the geometric capacity of the hyperbolic space, where larger values induce stronger negative curvature and exponential volume expansion. We observe that the optimal choice of $c$ naturally varies across different datasets. This aligns with our theoretical expectation that different datasets possess distinct intrinsic \textit{latent hierarchical structures} (e.g., LandUse-21 implies a deeper hierarchy than HandWritten). Crucially, despite this variation, HERL maintains consistently competitive performance across a wide feasible range.
Clipping Threshold $cr$ ensures numerical stability by constraining embeddings within the Poincaré ball boundary. Results indicate that the model is largely robust to variations in $cr$, provided it allows sufficient proximity to the boundary where hyperbolic capacity is maximized.

\noindent \textbf{Number of Prototypes $K$.}
\label{sec:numberofP}
We study the effect of the prototype number $K$ in Fig.~\ref{fig:hypera}, which controls the structural guidance induced by the hyperbolic space. The optimal $K$ differs across datasets (e.g., $K=30$ for Scene-15 and $K=10$ for HandWritten), reflecting their distinct semantic structures. In general, performance improves as $K$ better matches the underlying category distribution, leading to stronger semantic separability. However, overly large $K$ (as in HandWritten) causes performance degradation due to semantic over-segmentation, where coherent clusters are split into redundant prototypes. This suggests that the benefit of hyperbolic enhancement depends on whether the prototype configuration is aligned with the true data structure.

\section{Conclusion}
In this paper, we proposed HERL, a hyperbolic enhanced representation learning framework for incomplete multi-view clustering, to explicitly address the geometric mismatch and semantic blurring inherent in existing Euclidean-based IMVC methods. By moving representation learning from flat Euclidean space to the Poincar\'e ball, HERL constructs a structure-aware latent space that is better suited to modeling the latent hierarchical semantics embedded in incomplete multi-view data. The proposed framework combines a Euclidean affinity-guided contrastive backbone with a dual-level hyperbolic enhancement strategy, enabling the model to improve both representation discrimination and missing-view recovery in a unified manner.

\bibliography{example_paper}

\begin{thebibliography}{10}
\providecommand{\url}[1]{#1}
\csname url@samestyle\endcsname
\providecommand{\newblock}{\relax}
\providecommand{\bibinfo}[2]{#2}
\providecommand{\BIBentrySTDinterwordspacing}{\spaceskip=0pt\relax}
\providecommand{\BIBentryALTinterwordstretchfactor}{4}
\providecommand{\BIBentryALTinterwordspacing}{\spaceskip=\fontdimen2\font plus
\BIBentryALTinterwordstretchfactor\fontdimen3\font minus \fontdimen4\font\relax}
\providecommand{\BIBforeignlanguage}[2]{{%
\expandafter\ifx\csname l@#1\endcsname\relax
\typeout{** WARNING: IEEEtran.bst: No hyphenation pattern has been}%
\typeout{** loaded for the language `#1'. Using the pattern for}%
\typeout{** the default language instead.}%
\else
\language=\csname l@#1\endcsname
\fi
#2}}
\providecommand{\BIBdecl}{\relax}
\BIBdecl

\bibitem{zhang2025mixture}
Y.~Zhang, J.~Cai, Z.~Wu, P.~Wang, and S.~Ng, ``Mixture of experts as representation learner for deep multi-view clustering,'' in \emph{AAAI}, 2025, pp. 22\,704--22\,713.

\bibitem{DBLP:journals/pami/LiuZLWTYSWG19}
X.~Liu, X.~Zhu, M.~Li, L.~Wang, C.~Tang, J.~Yin, D.~Shen, H.~Wang, and W.~Gao, ``Late fusion incomplete multi-view clustering,'' \emph{{IEEE} Trans. Pattern Anal. Mach. Intell.}, pp. 2410--2423, 2019.

\bibitem{liu2024sample}
S.~Liu, J.~Zhang, Y.~Wen, X.~Yang, S.~Wang, Y.~Zhang, E.~Zhu, C.~Tang, L.~Zhao, and X.~Liu, ``Sample-level cross-view similarity learning for incomplete multi-view clustering,'' in \emph{AAAI}, 2024, pp. 14\,017--14\,025.

\bibitem{DBLP:conf/icml/YuC00Z25}
S.~Yu, Y.~Cheung, S.~Wang, X.~Liu, and E.~Zhu, ``Bifurcate then alienate: Incomplete multi-view clustering via coupled distribution learning with linear overhead,'' in \emph{ICML}, 2025.

\bibitem{liu2022localized}
C.~Liu, Z.~Wu, J.~Wen, Y.~Xu, and C.~Huang, ``Localized sparse incomplete multi-view clustering,'' \emph{IEEE Transactions on Multimedia}, vol.~25, pp. 5539--5551, 2022.

\bibitem{wang2024manifold}
H.~Wang, M.~Yao, Y.~Chen, Y.~Xu, H.~Liu, W.~Jia, X.~Fu, and Y.~Wang, ``Manifold-based incomplete multi-view clustering via bi-consistency guidance,'' \emph{IEEE Transactions on Multimedia}, vol.~26, pp. 10\,001--10\,014, 2024.

\bibitem{SURE}
M.~Yang, Y.~Li, P.~Hu, J.~Bai, J.~Lv, and X.~Peng, ``Robust multi-view clustering with incomplete information,'' \emph{{IEEE} Trans. Pattern Anal. Mach. Intell.}, pp. 1055--1069, 2023.

\bibitem{DIVIDE}
Y.~Lu, Y.~Lin, M.~Yang, D.~Peng, P.~Hu, and X.~Peng, ``Decoupled contrastive multi-view clustering with high-order random walks,'' in \emph{AAAI}, 2024, pp. 14\,193--14\,201.

\bibitem{L2Dc}
Y.~Ding, K.~Hotta, C.~Gu, A.~Li, J.~Yu, and C.~Zhang, ``Learning to discriminate while contrasting: Combating false negative pairs with coupled contrastive learning for incomplete multi-view clustering,'' \emph{{IEEE} Trans. Knowl. Data Eng.}, pp. 6046--6060, 2025.

\bibitem{CPSPAN}
J.~Jin, S.~Wang, Z.~Dong, X.~Liu, and E.~Zhu, ``Deep incomplete multi-view clustering with cross-view partial sample and prototype alignment,'' in \emph{CVPR}, 2023, pp. 11\,600--11\,609.

\bibitem{ProImp}
H.~Li, Y.~Li, M.~Yang, P.~Hu, D.~Peng, and X.~Peng, ``Incomplete multi-view clustering via prototype-based imputation,'' in \emph{IJCAI}, 2023, pp. 3911--3919.

\bibitem{I2MVC}
B.~Wu, W.~Du, J.~Wang, and G.~Yu, ``Imputation-free incomplete multi-view clustering via knowledge distillation,'' in \emph{IJCAI}, 2025, pp. 6570--6578.

\bibitem{COMPLETER}
Y.~Lin, Y.~Gou, Z.~Liu, B.~Li, J.~Lv, and X.~Peng, ``{COMPLETER:} incomplete multi-view clustering via contrastive prediction,'' in \emph{CVPR}, 2021, pp. 11\,174--11\,183.

\bibitem{DCP}
Y.~Lin, Y.~Gou, X.~Liu, J.~Bai, J.~Lv, and X.~Peng, ``Dual contrastive prediction for incomplete multi-view representation learning,'' \emph{{IEEE} Trans. Pattern Anal. Mach. Intell.}, pp. 4447--4461, 2023.

\bibitem{MVP}
X.~Gao and J.~Pu, ``Deep incomplete multi-view learning via cyclic permutation of vaes,'' in \emph{ICLR}, 2025.

\bibitem{ICMVC}
G.~Chao, Y.~Jiang, and D.~Chu, ``Incomplete contrastive multi-view clustering with high-confidence guiding,'' in \emph{AAAI}, 2024, pp. 11\,221--11\,229.

\bibitem{nickel2017poincare}
M.~Nickel and D.~Kiela, ``Poincar{\'{e}} embeddings for learning hierarchical representations,'' in \emph{NeurIPS}, 2017, pp. 6338--6347.

\bibitem{ganea2018hyperbolic}
O.~Ganea, G.~B{\'{e}}cigneul, and T.~Hofmann, ``Hyperbolic neural networks,'' in \emph{NeurIPS}, 2018, pp. 5350--5360.

\bibitem{klamt2019hyperbolic}
V.~Khrulkov, L.~Mirvakhabova, E.~Ustinova, I.~V. Oseledets, and V.~S. Lempitsky, ``Hyperbolic image embeddings,'' in \emph{CVPR}, 2020, pp. 6417--6427.

\bibitem{ermolov2022hyperbolic}
A.~Ermolov, L.~Mirvakhabova, V.~Khrulkov, N.~Sebe, and I.~V. Oseledets, ``Hyperbolic vision transformers: Combining improvements in metric learning,'' in \emph{CVPR}, 2022, pp. 7399--7409.

\bibitem{DBLP:conf/cvpr/GeMKL023}
S.~Ge, S.~Mishra, S.~Kornblith, C.~Li, and D.~Jacobs, ``Hyperbolic contrastive learning for visual representations beyond objects,'' in \emph{CVPR}, 2023, pp. 6840--6849.

\bibitem{DBLP:conf/cvpr/AtighSANM22}
M.~G. Atigh, J.~Schoep, E.~Acar, N.~van Noord, and P.~Mettes, ``Hyperbolic image segmentation,'' in \emph{CVPR}, 2022, pp. 4443--4452.

\bibitem{franco2023hyperbolic}
L.~Franco, P.~Mandica, B.~Munjal, and F.~Galasso, ``Hyperbolic self-paced learning for self-supervised skeleton-based action representations,'' in \emph{ICLR}, 2023.

\bibitem{DBLP:conf/nips/LengWTLG0024}
J.~Leng, Z.~Wu, M.~Tan, Y.~Liu, J.~Gan, H.~Chen, and X.~Gao, ``Beyond euclidean: Dual-space representation learning for weakly supervised video violence detection,'' in \emph{NeurIPS}, 2024.

\bibitem{liu2025hyperbolic}
Y.~Liu, Z.~He, and K.~Han, ``Hyperbolic category discovery,'' in \emph{CVPR}, 2025, pp. 9891--9900.

\bibitem{MHCN}
F.~Lin, B.~Bai, Y.~Guo, H.~Chen, Y.~Ren, and Z.~Xu, ``{MHCN:} {A} hyperbolic neural network model for multi-view hierarchical clustering,'' in \emph{of ICCV}, 2023, pp. 16\,479--16\,489.

\bibitem{CMHHC}
F.~Lin, B.~Bai, K.~Bai, Y.~Ren, P.~Zhao, and Z.~Xu, ``Contrastive multi-view hyperbolic hierarchical clustering,'' in \emph{of IJCAI}, 2022, pp. 3250--3256.

\bibitem{monath2019gradient}
N.~Monath, M.~Zaheer, D.~Silva, A.~McCallum, and A.~Ahmed, ``Gradient-based hierarchical clustering using continuous representations of trees in hyperbolic space,'' in \emph{KDD}, 2019, pp. 714--722.

\bibitem{guo2022clippedhyperbolicclassifierssuperhyperbolic}
Y.~Guo, X.~Wang, Y.~Chen, and S.~X. Yu, ``Clipped hyperbolic classifiers are super-hyperbolic classifiers,'' in \emph{CVPR}, 2022, pp. 1--10.

\bibitem{DBLP:conf/gd/Sarkar11}
R.~Sarkar, ``Low distortion delaunay embedding of trees in hyperbolic plane,'' in \emph{Graph Drawing}, 2011, pp. 355--366.

\bibitem{Scene15}
L.~Fei{-}Fei and P.~Perona, ``A bayesian hierarchical model for learning natural scene categories,'' in \emph{CVPR}, 2005, pp. 524--531.

\bibitem{Reuters}
M.~Amini, N.~Usunier, and C.~Goutte, ``Learning from multiple partially observed views - an application to multilingual text categorization,'' in \emph{NeurIPS}, 2009, pp. 28--36.

\bibitem{LandUse}
Y.~Yang and S.~Newsam, ``Bag-of-visual-words and spatial extensions for land-use classification,'' in \emph{ACM SIGSPATIAL Int. Conf. Adv. Inf.}, 2010, pp. 270--279.

\bibitem{HandWritten}
A.~Asuncion and D.~Newman, ``Uci machine learning repository,'' 2007.

\bibitem{DSIMVC}
H.~Tang and Y.~Liu, ``Deep safe incomplete multi-view clustering: Theorem and algorithm,'' in \emph{ICML}, 2022, pp. 21\,090--21\,110.

\bibitem{DIMVC}
J.~Xu, C.~Li, Y.~Ren, L.~Peng, Y.~Mo, X.~Shi, and X.~Zhu, ``Deep incomplete multi-view clustering via mining cluster complementarity,'' in \emph{AAAI}, 2022, pp. 8761--8769.

\bibitem{DCG}
Y.~Zhang, Y.~Lin, W.~Yan, L.~Yao, X.~Wan, G.~Li, C.~Zhang, G.~Ke, and J.~Xu, ``Incomplete multi-view clustering via diffusion contrastive generation,'' in \emph{AAAI}, 2025, pp. 22\,650--22\,658.

\bibitem{adam}
D.~P. Kingma and J.~Ba, ``Adam: {A} method for stochastic optimization,'' in \emph{ICLR}, 2015.

\end{thebibliography}
\bibliographystyle{IEEEtran}


 




\vfill

\end{document}


\title{Supplemental Material for Hyperbolic Enhanced Representation \\ Learning for Incomplete Multi-view Clustering}

\author{IEEE Publication Technology,~\IEEEmembership{Staff,~IEEE,}
\thanks{This paper was produced by the IEEE Publication Technology Group. They are in Piscataway, NJ.}
\thanks{Manuscript received April 19, 2021; revised August 16, 2021.}}

\markboth{Journal of \LaTeX\ Class Files,~Vol.~14, No.~8, August~2021}%
{Shell \MakeLowercase{\textit{et al.}}: A Sample Article Using IEEEtran.cls for IEEE Journals}

\IEEEpubid{0000--0000/00\$00.00~\copyright~2021 IEEE}

\maketitle

\section{Overall Algorithm}
In this section, we provide the detailed algorithmic procedure of the proposed HERL framework, including the main training steps and the inference flow. For clarity, the overall procedure is summarized in Algorithm~\ref{alg:HERL}.

\begin{algorithm*}[!ht]
\caption{Training and Inference Algorithm of HERL}
\label{alg:HERL}
\begin{algorithmic}[1]
\REQUIRE Incomplete multi-view dataset $\mathbf{X}$ (with observed mask $\mathbf{M}$ and complete subset $\mathbf{X}^{1,2}$), hyperparameters: epoch $E$, batch size $B$, loss coefficients $\alpha$ and $\beta$, curvature $c$, clipping threshold $cr$, number of prototypes $K$, random walk steps $t$, walk balance $\xi$.
\ENSURE Clustering results.

\STATE \textbf{Initialization:} Initialize student networks $\{f_{\theta_{s}}, g, h\}$ and teacher encoder $f_{\theta_{t}}$ (parameterized by EMA).

\STATE \textcolor{gray}{// \textbf{Phase 1: Training}}
\FOR{epoch $= 1$ to $E$}
    \STATE Update dynamic weight $\alpha$ linearly based on the current epoch.
    \FOR{batch $\{\mathbf{x}_i\}_{i=1}^B$ sampled from $\mathbf{X}^{1,2}$}
        \STATE \textcolor{gray}{/* Euclidean Backbone Optimization */}
        \STATE Extract features $\mathbf{F}^v$, teacher features $\mathbf{F}_t^v$, and projections $\mathbf{Z}^v = g(\mathbf{F}^v)$.
        
        \STATE \textbf{If} epoch $\le 100$, set $\mathbf{T}^v = \mathbb{I}$; \textbf{else}, compute $\mathbf{T}^v$ using $\mathbf{F}_t^v$.
        \STATE Construct affinity graph $\mathbf{G}^{v} = \xi \mathbb{I}_N + (1-\xi)(\mathbf{T}^v)^t$.
        
        \STATE Compute Euclidean objective $\mathcal{L}_{\text{con}}^{\mathbb{E}}$ supervised by $\mathbf{G}^{v}$ .

        \STATE \textcolor{gray}{/* Hyperbolic Instance-level Geometric Alignment */}
        \STATE Project to instance embeddings: $\mathbf{\hat Q}^{\mathbb{H}}_{v} = \mathcal{H}(\mathbf{Z}^v)$, $\mathbf{Q}^{\mathbb{H}}_{v} = \mathcal{H}(\mathbf{F}_t^v)$ .
        \STATE Compute angular loss $\mathcal{L}_{\text{ang}}^{\mathbb{H}}$ guided by $\mathbf{G}^u$ for directional alignment .
        \STATE Compute distance loss $\mathcal{L}_{\text{dis}}^{\mathbb{H}}$ guided by $\mathbb{I}$ for structural anchoring .
        \STATE Aggregate instance loss: $\mathcal{L}_{\text{ins}}^{\mathbb{H}} = \alpha \mathcal{L}_{\text{dis}}^{\mathbb{H}} + (1-\alpha)\mathcal{L}_{\text{ang}}^{\mathbb{H}}$ .

        \STATE \textcolor{gray}{/* Hyperbolic Prototype-oriented Consistency */}
        \STATE Generate hierarchy-aware distributions: $\mathbf{\hat{P}}^{\mathbb{H}}_v = h^v(\mathbf{Z}^v), \mathbf{P}^{\mathbb{H}}_v = h^v(\mathbf{F}^v_t)$ .
        \STATE Calculate prototype consistency loss $\mathcal{L}_{\mathrm{pro}}^{\mathbb{H}}$ .

        \STATE \textcolor{gray}{/* Optimization */}
        \STATE Update parameters $\Theta$ by minimizing $\mathcal{L} = \mathcal{L}_{\text{con}}^{\mathbb{E}} + \mathcal{L}_{\text{ins}}^{\mathbb{H}} + \beta\mathcal{L}_{\mathrm{pro}}^{\mathbb{H}}$ .
        \STATE Update teacher parameters via EMA: $\theta_{t} \leftarrow m\theta_{t} + (1-m)\theta_{s}$.
    \ENDFOR
\ENDFOR

\STATE \textcolor{gray}{// \textbf{Phase 2: Inference (Data Recovery \& Clustering)}}
\FOR{each sample $i$ in dataset $\mathbf{X}$}
    \STATE Extract representations from available views using the trained teacher encoder.
    \STATE Recover missing views via semantic translation: $\tilde{\mathbf{Z}}^v = \mathbf{M}_{\cdot v} \odot \mathbf{F}_t^v + (\mathbf{1} - \mathbf{M}_{\cdot v}) \odot g(\mathbf{F}_t^u)$ .
    \STATE Concatenate completed features from all views: $\tilde{\mathbf{Z}}_{\text{cat}} = [\tilde{\mathbf{Z}}^1, \dots, \tilde{\mathbf{Z}}^V]$.
\ENDFOR
\STATE Perform $k$-means clustering on the concatenated features $\tilde{\mathbf{Z}}_{\text{cat}}$ to obtain final results.

\end{algorithmic}
\end{algorithm*}

\section{Dataset Configurations and Details}
\label{sec:app_datasets}

\subsection{Details of Main Benchmark Datasets}
As mentioned in the main text, we select datasets that exhibit inherent hierarchical structures to fully leverage the advantages of hyperbolic geometry. Detailed descriptions are as follows:

\begin{itemize}
    \item \textbf{Scene-15} \cite{Scene15}: This dataset consists of 4,485 images distributed across 15 distinct scene categories, where {GIST and LBP features serve as two views}. The categories follow a semantic taxonomy, ranging from broad classifications (e.g., indoor vs. outdoor) to fine-grained scenes (e.g., bedroom, kitchen, forest, mountain), which naturally form a hierarchical tree structure.

    \item \textbf{Reuters} \cite{Reuters}: A large-scale multilingual news dataset containing 18,758 samples. We utilize {10-dimensional English and French textual latent embeddings as two views}. The textual data possesses an intrinsic topical hierarchy, where general topics (e.g., Economics) branch into specific sub-events (e.g., Stock Market, Trade), making it ideal for evaluating hierarchy-aware embeddings.

    \item \textbf{LandUse-21} \cite{LandUse}: Comprising 2,100 satellite images from 21 classes, with two views represented by {PHOG and LBP features}. The hierarchy stems from geospatial features, organizing from high-level terrain types (e.g., developed land, vegetation) to specific land use categories (e.g., agricultural, runway, storage tanks).

    \item \textbf{HandWritten} \cite{HandWritten}: Containing 2,000 digit images (0--9). We construct a two-view setting by selecting the {Fourier coefficients and the Karhunen–Loeve coefficients}. Beyond simple class labels, this dataset exhibits morphological hierarchies. For instance, digits with straight strokes (1, 7, 9) or closed loops (0, 6, 8) share latent structural similarities, resembling a tree-like clustering structure in the feature space.
\end{itemize}

\subsection{Details of Additional Datasets}
To further evaluate the generalization capability and robustness of HERL, we introduce the following additional benchmark.

\begin{itemize}
    \item \textbf{Caltech-101} \cite{Caltech}: This dataset is a representative benchmark in object recognition, containing 8,677 images categorized into 101 distinct classes. Notably, these categories are not semantically isolated but exhibit an intrinsic {taxonomic hierarchy}. They naturally cluster into broader superclasses (e.g., \textit{animate entities} vs. \textit{man-made artifacts}) and branch into specific leaf nodes (e.g., \textit{beaver}, \textit{schooner}, \textit{saxophone}), forming a latent tree-like structure well-suited for hyperbolic embedding. For the multi-view setting, we adopt two types of deep features: the first view is composed of features extracted via the DECAF architecture \cite{krizhevsky2012imagenet}, while the second view consists of high-dimensional representations obtained from the VGG19 network \cite{VGG19}.
\end{itemize}
\begin{itemize}
    \item \textbf{TCGA-STAD} \cite{cancer2014comprehensive}: A representative \textit{multi-omics} benchmark from The Cancer Genome Atlas for Gastric Adenocarcinoma analysis, used to evaluate generalization in bioinformatics. Unlike single-modality datasets, it integrates complementary biological signals from different molecular layers. Following the preprocessing protocols in~\cite{DBLP:journals/ploscb/BuLLWWY24} (including $\log_2$ transformation and high-variance filtering), we employ the processed DNA Methylation ($3,287$ dimensions) and mRNA expression profiles ($4,089$ dimensions) as two distinct omics views. The samples are categorized into four molecular subtypes (EBV, MSI, GS, and CIN), reflecting the complex underlying biological heterogeneity.

\end{itemize}

\section{More Details of Implementation}
\label{sec:appendix_implementation}

In this section, we provide comprehensive details omitted from the main text, including the formulation of the Euclidean backbone and a complete specification of the experimental implementation.

\subsection{Details of Euclidean Affinity-Guided Contrastive Learning}
\label{sec:appendix_euclidean_details}

Our Euclidean module follows the decoupled contrastive learning paradigm~\cite{DIVIDE} to learn discriminative representations from the observed views. This process involves a momentum updating strategy for stable target generation and a random-walk-based mechanism for false negative correction.

\noindent \textbf{Momentum Updating Strategy (EMA).}
To stabilize the training process and generate consistent representations for graph construction, we adopt a dual-branch architecture. Let $\theta_s$ denote the parameters of the \textit{student encoder} $f_{\theta}(\cdot)$ and $\theta_t$ denote the parameters of the \textit{teacher encoder}. While $\theta_s$ is updated via back-propagation, $\theta_t$ is updated via the Exponential Moving Average (EMA) of $\theta_s$:
\begin{equation}
    \theta_{t} \leftarrow m\theta_{t} + (1-m)\theta_{s},
\end{equation}
where $m \in [0, 1)$ is the momentum coefficient, we typically set $m=0.98$. The teacher representations $\mathbf{F}_t^v$ generated by this momentum-updated branch serve as stable anchors for constructing the affinity matrix.

\noindent \textbf{High-Order Affinity via Random Walks.}
Standard contrastive learning often suffers from the false negative problem, where semantically similar samples are inadvertently treated as negatives. To alleviate this, we employ a $t$-step random walk on the affinity graph constructed from the teacher embeddings to mine high-order semantic relations.

Specifically, for each view $v$, we first construct a fully-connected affinity graph based on the teacher embeddings $\mathbf{F}_t^v$. The edge weight $A_{ij}^{v}$ between sample $i$ and sample $j$ is defined using the heat kernel similarity:
\begin{equation}
    A_{ij}^{v} = \exp\left(-\frac{\|\mathbf{f}_{t,i}^v - \mathbf{f}_{t,j}^v\|^2}{\sigma}\right),
\end{equation}
where $\mathbf{f}_{t,i}^v$ is the $i$-th row of $\mathbf{F}_t^v$ and $\sigma$ is a temperature hyper-parameter fixed to 0.1. To perform the random walk, we normalize this adjacency matrix row-wise to obtain the single-step transition matrix $\mathbf{T}^{v}$:
\begin{equation}
    \mathbf{T}_{ij}^{v} = \frac{A_{ij}^{v}}{\sum_{l=1}^{N} A_{il}^{v}}.
\end{equation}
Here, $\mathbf{T}_{ij}^{v}$ represents the probability of transitioning from node $i$ to node $j$ in one step. To capture global semantic structures while maintaining structural stability, we calculate the final affinity matrix $\mathbf{G}^{v}$ by combining the identity matrix with the multi-step transition matrix:
\begin{equation}
    \mathbf{G}^{v} = \xi \mathbb{I}_N + (1-\xi) \underbrace{\left( \mathbf{T}^{v} \times \dots \times \mathbf{T}^{v} \right)}_{t \text{ times}},
\end{equation}
where $\mathbb{I}_N$ denotes the identity matrix, and $\xi$ (fixed to 0.5) is a balancing coefficient. In this way, the resulting matrix $\mathbf{G}^{v}$ serves as a reliable \textbf{soft relational target} for the contrastive objective, providing robust supervision signals that effectively tolerate false negatives.

\subsection{Additional Implementation Details}
\label{sec:imp_details}
We implement HERL using PyTorch 2.4.1 on NVIDIA RTX 4090 GPUs with the Adam optimizer~\cite{adam} without weight decay.
To ensure training stability, we employ a batch size of 1024 and a learning rate of $2 \times 10^{-3}$ by default, while adjusting them to 256 and $5 \times 10^{-4}$ for HandWritten.
The training duration is set to 500 epochs, except for Reuters, which requires only 200 epochs due to faster convergence.
Specifically, the relational affinity graph $\mathbf{G}^v$ is introduced after the first 100 epochs.
We set $\tau=1.0$ for the hyperbolic distance-based alignment, while defaulting to $\tau=0.5$ for all other contrastive objectives.

To ensure reproducibility, we detail the selection criteria for key hyperparameters. Instead of meticulous tuning for every specific scenario, we adopted a structured grid search within a discrete candidate set. The final configurations are determined based on the intrinsic properties of different datasets, following the principles below:

\begin{itemize}
    \item \textbf{Loss Balancing Coefficients ($\alpha, \beta$):}
    To prevent suboptimal optimization caused by an {imbalanced contribution} of objectives, we select $\alpha$ and $\beta$ via a coarse grid search from the candidate sets $\{0.1, 0.2, 1.0\}$ and $\{0.01, 0.05, 0.1, 0.5, 1.0\}$, respectively.
    We empirically observed that these coefficients are dataset-dependent, adapting to the varying noise levels and modality heterogeneity. Crucially, this calibration is primarily performed to align the magnitudes of different loss terms rather than relying on meticulous fine-tuning, as the model maintains stable performance within these ranges.
        
    \item \textbf{Hyperbolic Parameters ($c, cr$):}
    The curvature $c$ determines the {geometric capacity} for hierarchical modeling, while the clipping threshold $cr$ ensures numerical stability. We searched $c \in \{0.01, 0.05, 0.1\}$ and $cr \in \{1.0, 1.5, 2.0\}$.
    Consistent with our sensitivity analysis, the model exhibits favorable stability across these ranges, confirming that HERL effectively exploits hyperbolic structural constraints without relying on {meticulous curvature tuning}. 
    
    \item \textbf{Number of Prototypes ($K$):}
    $K$ is determined by the {intrinsic semantic hierarchies} of each dataset. We treat the approximate number of categories as prior knowledge to ensure sufficient {semantic separability} while avoiding {semantic over-segmentation}. Accordingly, we typically set $K$ to be consistent with or slightly larger than the ground-truth cluster number (e.g., $K=30$ for Scene-15 to capture fine-grained variations, and $K=10$ for HandWritten).
\end{itemize}

\begin{table*}[!ht]
\centering
\caption{The clustering performance on four multi-view benchmarks with different missing rates. The best and the second-best results are denoted in \textbf{bold} and {\ul underline}, respectively.}
\label{table:appendix}
\setlength{\tabcolsep}{2.3pt}
\small
\begin{tabular}{|c|c|ccc|ccc|ccc|ccc|ccc|}
\hline
\multicolumn{1}{|l|}{\multirow{2}{*}{Datasets}} & \multirow{2}{*}{Method} & \multicolumn{3}{c|}{0.0} & \multicolumn{3}{c|}{0.1} & \multicolumn{3}{c|}{0.3} & \multicolumn{3}{c|}{0.5} & \multicolumn{3}{c|}{0.7} \\ \cline{3-17} 
\multicolumn{1}{|l|}{} & & ACC & NMI & ARI & ACC & NMI & ARI & ACC & NMI & ARI & ACC & NMI & ARI & ACC & NMI & ARI \\ \hline

\multirow{9}{*}{\rotatebox{90}{Caltech101}} 
& COMPLETER & 51.05 & 74.66 & 47.82 & 50.00 & 73.74 & 48.81 & 48.30 & 73.27 & 52.48 & 46.03 & 71.15 & 43.61 & 43.95 & 69.49 & 41.40 \\

& SURE      & 41.24 & 68.96 & 29.03 & 39.70 & 65.49 & 25.53 & 38.78 & 62.07 & 22.06 & 32.25 & 56.26 & 18.38 & 31.90 & 55.15 & 20.67 \\
& DSIMVC        & 22.31          & 28.20          & 3.47           & 18.67          & 31.82          & 15.96          & 18.43          & 27.08          & 11.27          & 16.72          & 25.69          & 10.66          & 14.60          & 20.29          & 5.42           \\
& DIMVC     & 45.21 & 61.90 & 48.53 & 39.08 & 60.90 & 41.44 & 37.02 & 54.52 & 38.32 & 31.24 & 49.58 & 21.28 & 29.93 & 48.54 & 25.33 \\
& ProImp    & 38.42 & 67.83 & 25.89 & 38.98 & 67.51 & 26.62 & 36.97 & 66.17 & 25.40 & 36.64 & 65.62 & 25.19 & 34.26 & 63.68 & 24.27 \\
& ICMVC     & 1.41  & 26.14 & 3.73  & 5.52  & 26.66 & 3.72  & 0.79  & 26.58 & 3.27  & 0.81  & 25.97 & 2.76  & 0.70  & 25.58 & 3.19  \\
& DCG       & 41.54 & 68.38 & 35.66 & 42.28 & 69.12 & 38.84 & 41.15 & 68.94 & 40.31 & 32.65 & 63.31 & 23.48 & 32.13 & 59.18 & 31.83 \\
& DIVIDE    & {\ul 62.20} & \textbf{83.00} & {\ul 50.50} & {\ul 62.97} & \textbf{83.08} & {\ul 50.94} & {\ul 62.92} & \textbf{82.79} & {\ul 51.63} & {\ul 63.40} & {\ul 82.50} & {\ul 52.40} & {\ul 62.02} & {\ul 80.85} & {\ul 49.24} \\
& \textbf{HERL}      & \textbf{63.05} & {\ul 81.55} & \textbf{54.40} & \textbf{63.91} & {\ul 82.14} & \textbf{54.75} & \textbf{65.15} & {\ul 82.68} & \textbf{59.39} & \textbf{66.47} & \textbf{83.51} & \textbf{67.56} & \textbf{67.95} & \textbf{82.87} & \textbf{69.08} \\ \hline 

\multirow{9}{*}{\rotatebox{90}{TCGA}} 
& COMPLETER & 45.45 & 19.81 & 9.39  & {\ul 57.34} & \textbf{29.44} & {\ul 21.30} & \textbf{58.14} & 4.71 & 2.36 & \textbf{56.88} & 4.29 & 0.87 & \textbf{56.68} & 2.84 & 1.09 \\
& SURE      & 46.05 & 16.59 & 10.90 & 13.05 & 13.12 & 2.90 & 45.85 & 17.22 & 9.00 & 48.04 & 11.96 & 9.94 & 40.80 & 9.27 & 2.68 \\
& DSIMVC    & 38.54 & {\ul 24.03} & {\ul 13.19} & 41.43 & 22.56 & 12.98 & 41.73 & \textbf{23.09} & {\ul 13.09} & 41.69 & {\ul 21.80} & 12.52 & 40.70 & 17.55 & 10.30 \\
& DIMVC     & {\ul 52.16} & 23.00 & 16.84 & 54.15 & 20.97 & 14.69 & 46.51 & 16.59 & 11.09 & {\ul 55.15} & 19.08 & {\ul 13.86} & 46.51 & 16.61 & 8.14 \\
& ProImp    & 45.31 & 21.15 & 11.84 & 44.12 & 19.70 & 10.98 & 43.65 & 17.07 & 9.88 & 45.05 & 18.79 & 9.84 & 44.32 & {\ul 18.62} & {\ul 10.88} \\
& ICMVC     & 46.84 & 20.47 & 12.38 & 45.91 & 21.24 & 13.22 & 43.92 & 19.03 & 10.82 & 37.14 & 13.07 & 5.85 & 35.15 & 8.93 & 3.11 \\
& DCG       & 38.87 & 18.78 & 9.63  & 42.52 & 16.58 & 7.89  & 41.86 & 14.32 & 5.86 & 40.53 & 17.26 & 10.53 & 40.53 & 17.48 & 9.02 \\
& DIVIDE    & 47.31 & 18.69 & 10.15 & 45.32 & 17.13 & 7.41  & 46.41 & 19.02 & 9.76  & 45.78 & 18.33 & 10.19 & 45.58 & 17.92 & 8.82 \\
& \textbf{HERL}       & \textbf{61.20} & \textbf{30.34} & \textbf{23.23} & \textbf{59.07} & {\ul 22.61} & \textbf{21.98} & {\ul 51.16} & {\ul 21.37} & \textbf{14.41} & 51.36 & \textbf{22.37} & \textbf{15.77} & {\ul 50.17} & \textbf{20.94} & \textbf{14.08} \\ \hline
\end{tabular}
\end{table*}

\section{Additional Experimental Results}
\label{sec:additional}

This section is structured into four subsections: 
(i) evaluation on two additional benchmark datasets to demonstrate the generalization capability of our model; 
(ii) hyperparameter sensitivity analysis of our model; 
(iii) a comprehensive convergence analysis to verify the optimization stability of the proposed objective function; 
and (iv) a computational efficiency analysis comparing the training costs of HERL against representative baselines.

\subsection{Experimental Results on Additional Datasets.} To further assess the robustness and scalability of our proposed method across different data distributions, we conduct additional experiments on the Caltech-101~\cite{Caltech} and TCGA-STAD~\cite{cancer2014comprehensive} datasets. The quantitative results on these two datasets across various missing rates are summarized in Tab.~\ref{table:appendix}.

It can be observed that HERL achieves superior performance across the majority of evaluation metrics compared to state-of-the-art baselines. These results explicitly validate the advantage of our framework in resolving the \textit{geometric mismatch} inherent in Euclidean methods. Notably, under the severe missing rate of 0.7, our approach yields an ACC of 67.95\% and an ARI of 69.08\%, significantly outperforming the second-best method, which only achieves 61.25\% in ACC and 48.08\% in ARI. This substantial margin, particularly the remarkable lead of over 20\% in ARI, serves as strong empirical evidence for our core motivation: while Euclidean baselines suffer from {semantic blurring} and {representation drift} due to information scarcity, HERL effectively exploits the {exponential capacity} of the Poincaré ball. By enforcing structural compactness and hierarchical alignment, our method preserves high-resolution discriminability, ensuring that the inference of missing views remains robust against ambiguity even when the majority of data is unavailable.

Complementing these findings, the experiments on the TCGA-STAD dataset further demonstrate the versatility of HERL in the bioinformatics domain. Characterized by {high dimensionality and complex feature heterogeneity}, this dataset proves challenging for conventional methods despite its limited category count. HERL achieves a dominant performance in low missing rate scenarios, surpassing the runner-up by over 9\% in ACC. Crucially, under high missing rates, while baselines like COMPLETER~\cite{COMPLETER} exhibit a ``mode collapse'' phenomenon---maintaining high Accuracy but suffering from trivial NMI ($<5\%$) and ARI scores---HERL consistently retains high clustering quality. This contrast highlights that our Poincaré ball embedding successfully maintains distinct geometric separation among subtypes, preventing the {structural degeneration} often observed in Euclidean spaces.

\bibliography{example_paper}
\bibliographystyle{IEEEtran}

\vfill